\documentclass[10pt,twocolumn,letterpaper]{article}

\usepackage{cvpr}              %

\renewcommand{\paragraph}[1]{\vspace{.2em}\noindent\textbf{#1}}

\definecolor{cvprblue}{rgb}{0.21,0.49,0.74}
\usepackage[table]{xcolor}
\usepackage[pagebackref,breaklinks,colorlinks,allcolors=cvprblue]{hyperref}
\usepackage{multirow}
\usepackage{xcolor}   %
\usepackage{pifont}   %
\usepackage{placeins}
\def\paperID{} %
\def\confName{CVPR}
\def\confYear{2026}

\newcommand{\asia}[1]{\textcolor{black}{#1}}

\newcommand{\marek}[1]{\textcolor{black}{#1}}
\usepackage{pifont}
\newcommand{\ours}{\textbf{LumiMotion}}

\title{LumiMotion: Improving Gaussian Relighting with Scene Dynamics}

\author{
Joanna Kaleta$^{1,2}$\thanks{Equal contribution}, \quad
Piotr Wójcik$^{3,4,5}$\footnotemark[1], \quad
Kacper Marzol$^{6}$ \\
Tomasz Trzciński$^{1,7}$, \quad
Kacper Kania$^{1}$, \quad
Marek Kowalski$^{8}$ \\
\\
$^{1}$Warsaw University of Technology \quad
$^{2}$Sano Centre for Computational Medicine\\
$^{3}$Inst. for Biomedical Informatics, Univ. Hospital of Cologne \quad
$^{4}$University of Cologne\\
$^{5}$CMMC Cologne \quad
$^{6}$Jagiellonian University \quad
$^{7}$IDEAS Research Institute \quad
$^{8}$Microsoft
}
\begin{document}
\maketitle
\vspace{-0.7em}
\begin{abstract}
In 3D reconstruction, the problem of inverse rendering, namely recovering the illumination of the scene and the material properties, is fundamental. Existing Gaussian Splatting-based methods primarily target static scenes and often assume simplified or moderate lighting to avoid entangling shadows with surface appearance. This limits their ability to accurately separate lighting effects from material properties, particularly in real-world conditions. We address this limitation by leveraging dynamic elements---regions of the scene that undergo motion---as a supervisory signal for inverse rendering. Motion reveals the same surfaces under varying lighting conditions, providing stronger cues for disentangling material and illumination. This thesis is supported by our experimental results which show we improve LPIPS by 23\% for albedo estimation and by 15\% for scene relighting relative to next-best baseline. To this end, we introduce \ours, the first Gaussian-based approach that leverages dynamics for inverse rendering and operates in arbitrary dynamic scenes. Our method learns a dynamic 2D Gaussian Splatting representation that employs a set of novel constraints which encourage the dynamic regions of the scene to deform, while keeping static regions stable. As we demonstrate, this separation is crucial for correct optimization of the albedo. Finally, we release a new synthetic benchmark comprising five scenes under four lighting conditions, each in both static and dynamic variants, for the first time enabling systematic evaluation of inverse rendering methods in dynamic environments and challenging lighting. Link to project page in footnote\footnote{\url{https://joaxkal.github.io/LumiMotion/}}.

\end{abstract}
\vspace{-0.6cm}

\section{Introduction}

\begin{figure}[!tbhp]
\centering
\includegraphics[width=\linewidth]{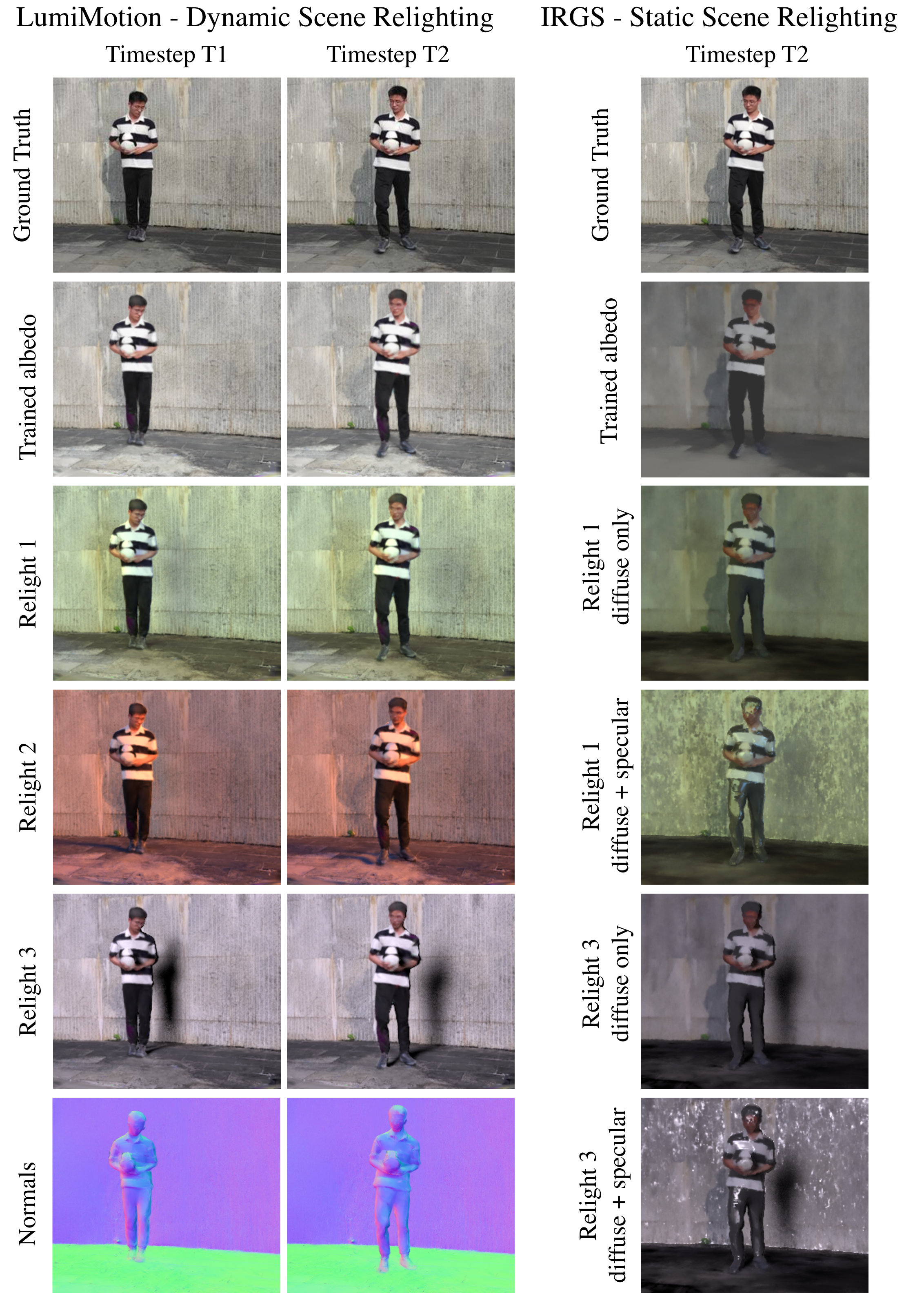}
\vspace{-1.5em}
\caption{\textbf{Qualitative results for \ours{} and IRGS \cite{irgs} on a real-world dataset.} Note that thanks to modeling the scene dynamics \ours{} can successfully remove shadows from the albedo, leading to improved relighting. IRGS struggles to correctly remove shadows from the albedo, and the specular component is not properly optimized. To allow for comparison of \ours{} and IRGS in absence of the latter's specularity estimation errors we show some of IRGS results with diffuse component only.}
\label{fig:real_world_qual}
\vspace{-1.9em}
\end{figure}
Inverse rendering - the task of recovering geometry, material properties, and illumination from a set of images of a scene - is a fundamental challenge in computer vision and graphics \cite{first}. Although methods like Neural Radiance Fields \cite{nerf} and Gaussian Splatting \cite{3d_gaussian} allow for accurately representing scene geometry, they represent the color of the reconstructed scene as it was observed, including shadows and other illumination-dependent effects.
This conflation of lighting and material means the scene cannot be rendered under a different illumination (it cannot be relit). This limits the applicability of those reconstructions in fields such as gaming or film making, which require control over illumination. Recent works have introduced physically-motivated decomposition methods to address this limitation \cite{irgs, GI-GS, R3DG2023}, though, as shown in~\cref{fig:real_world_qual}, such approaches may struggle in real-world scenes with significant direct illumination.

We hypothesize that the reason existing methods struggle is that they operate on static scenes, where the content is not moving. This lack of motion means that it is difficult to discern between the intrinsic color of objects and the observed color, which is a function of the incident light. For example, when viewing images of static scenes, it may be difficult to determine whether a certain part of the scene is darker because a shadow is cast on it or because the material itself is dark. Thus, we propose \ours, the first Gaussian-based method to perform inverse rendering of arbitrary dynamic scenes. Based solely on a video sequence of a scene with dynamic elements, our approach reconstructs the scene’s geometry, as well as its material properties and illumination. The lighting conditions are assumed to be static, not known a priori and are inferred as part of the training. We leverage the scene dynamics to improve the quality of the estimated illumination and albedo, thereby enhancing the ability to render those scenes under novel illumination conditions. 

\ours~operates in two stages. In the first stage, we jointly learn the static scene geometry and a neural network that models the dynamics of the scene including changes to shape and color of objects in the scene. The geometry is modeled with 2D Gaussian Splats \cite{2DG} that represent the surfaces in the scene with a collection of flat disks. A key benefit of this approach over 3D Gaussian Splatting or Neural Radiance Fields is that the 2D Gaussians have normal vectors associated with them and those are necessary for our second stage. In the second stage we freeze both the geometry and the neural network and proceed to inverse rendering. At this point we move away from rendering the Gaussian color directly and instead we compute the color of each Gaussian as a function of its material and the incident light that is computed via ray tracing. This allows for jointly optimizing the material (described jointly by albedo and roughness) and the illumination. 

Once we have those parameters, we can relight the scene with novel illumination or use the estimated illumination to relight other scenes. We demonstrate those abilities qualitatively on real data and quantitatively on a new synthetic dataset we created, where the ground truth illumination and material are known. Please see~\cref{fig:real_world_qual} for a teaser of the results.

\noindent Our contributions can be summarized as follows:
\begin{itemize}
    \item First Gaussian-based approach for inverse rendering in arbitrary dynamic scenes, achieving better separation of material and illumination by leveraging scene dynamics as a supervisory signal.
    \item A novel set of constraints on the deformation network that allows for better separation of static and dynamic parts of the scene and for improved modeling of scene geometry in time.
    \item A new synthetic dataset allowing for comparing the performance of inverse rendering approaches for static and dynamic scenes.
\end{itemize}
\newcommand{\cmark}{\textcolor{green!80!black}{\ding{51}}} %
\newcommand{\xmark}{\textcolor{red!80!black}{\ding{55}}}   %

\begin{table}[!tb]
\centering
\caption{\textbf{Overview of the representative relightable methods and datasets.} Existing methods often target static scenes or focus on human avatars, assume known lights or require unnatural lighting setup. In contrast, \ours~ uniquely performs inverse rendering of generic dynamic scenes under unknown natural lighting. We also show that \textbf{none} of the available datasets with known GT lighting meet the requirements for the tackled setup, which we address with our newly released dataset.}

\resizebox{\linewidth}{!}{
\begingroup
\fontsize{14pt}{14pt}\selectfont
\begin{tabular}{l
  >{\centering\arraybackslash}p{2.3cm}
  >{\centering\arraybackslash}p{2.5cm}
  >{\centering\arraybackslash}p{2.5cm}
  >{\centering\arraybackslash}p{2.5cm}}
\toprule
\multicolumn{5}{c}{\textbf{Relightable methods}} \\
\midrule
Method Name &
Supports Dynamics &
Object-Agnostic &
Unknown Train Light &
Natural Train Light \\
\midrule
GS$^{3}$~\cite{GS³}  & \xmark & \cmark & \xmark & \xmark \\
R-3DGS~\cite{R3DG2023}, GI-GS~\cite{GI-GS}, IR-GS~\cite{irgs} & \xmark & \cmark & \cmark & \cmark \\
TensorIR~\cite{TensorIR}, NeRFactor~\cite{nerfactor} 
& \xmark & \cmark & \cmark & \cmark \\
Relightable Neural Actor~\cite{luvizonGKHT24} & \cmark & \xmark & \xmark & \cmark \\
Gaussian Codec Avatars~\cite{codec_avatars} \&~\cite{WangARXIV2025_relightable_full_body} & \cmark & \xmark & \xmark & \xmark \\
IntrinsicAvatar~\cite{intrinistic_avatars}  & \cmark & \xmark & \cmark & \cmark \\
Relightable [...] Neural Avatars~\cite{zhan2024interactive_rel_human_avatars}  & \cmark & \xmark & \cmark & \cmark \\
\ours & \cmark & \cmark & \cmark & \cmark \\

\bottomrule
\end{tabular}
\endgroup
}
\vspace{0.8em}

\resizebox{\linewidth}{!}{%
\begingroup
\fontsize{14pt}{14pt}\selectfont
\begin{tabular}{l
  >{\centering\arraybackslash}p{2.3cm} %
  >{\centering\arraybackslash}p{2.4cm} %
  >{\centering\arraybackslash}p{2.6cm} %
  >{\centering\arraybackslash}p{3.5cm}} %
\toprule
\multicolumn{5}{c}{\textbf{Relightable datasets}} \\
\midrule
Dataset Name &
Domain &
Scene Type &
Train Light Type &
Static-Dynamic Reference \\
\midrule
OLAT~\cite{liu2023openillumination} & \textcolor{green!80!black}{generic} & \textcolor{red!80!black}{static}  & \textcolor{red!80!black}{OLAT} & \xmark \\
TensoIR~\cite{TensorIR}              & \textcolor{green!80!black}{generic} & \textcolor{red!80!black}{static}  & \textcolor{green!80!black}{natural} & \xmark \\
Synthetic4Relight~\cite{synthetic4relight} & \textcolor{green!80!black}{generic} & \textcolor{red!80!black}{static} & \textcolor{green!80!black}{natural} & \xmark \\
Stanford-ORB~\cite{kuang2023stanfordorb} & \textcolor{green!80!black}{generic} & \textcolor{red!80!black}{static} & \textcolor{green!80!black}{natural} & \xmark \\
RANA~\cite{rana} & \textcolor{red!80!black}{avatars} & \textcolor{green!80!black}{dynamic}  & \textcolor{green!80!black}{natural} & \xmark \\
Codec Avatar Studio~\cite{codec_avatar}& \textcolor{red!80!black}{avatars} & \textcolor{green!80!black}{both} & \textcolor{red!80!black}{multi-light} & \xmark \\
\ours & \textcolor{green!80!black}{generic} & \textcolor{green!80!black}{both} & \textcolor{green!80!black}{natural} & \cmark \\
\bottomrule
\end{tabular}
\endgroup
}
\vspace{-0.3cm}
\label{tab:methods_datasets}
\end{table}

\section{Related Work}

\paragraph{Novel View Synthesis.} Novel view synthesis aims to generate images of a scene from novel viewpoints using a limited set of input observations. Neural Radiance Fields (NeRF)~\cite{nerf} marked a breakthrough in novel view synthesis, offering high quality of view-dependent renderings. Despite their strengths, NeRFs suffer from slow training and rendering. Several approaches were made to overcome this limitation \cite{muller2022instantngp, garbin2021fastnerf, TensorRF}. Recently, 3D Gaussian Splatting \cite{3d_gaussian} proposed representing scenes with a gaussian point cloud rendered with a tile-based rasterizer, achieving state-of-the-art results with lower computational cost. A series of subsequent studies tackled a range of challenges, including editability \cite{gsedit, editor, deferred}, realistic modeling of conditioned appearance \cite{kaleta2025lumigauss, xie2024envgs}, modeling of motion and scene dynamics \cite{SC-GS, Yang_2024_CVPR_deformable_gs, Wu_2024_CVPR_4dgaussians, liu2024dynamic} or improved geometry reconstruction \cite{2DG, guedon2023sugar, Choi_2024_ACCV_meshgs}. Utilizing disc-like, flat Gaussian primitives in 2DGS~\cite{2DG} improved surface reconstruction quality while well-defined ray-splat intersection provided a straightforward foundation for extending the method to various tasks.

\begin{figure*}[!tbhp]
  \centering
  \includegraphics[width=\linewidth]{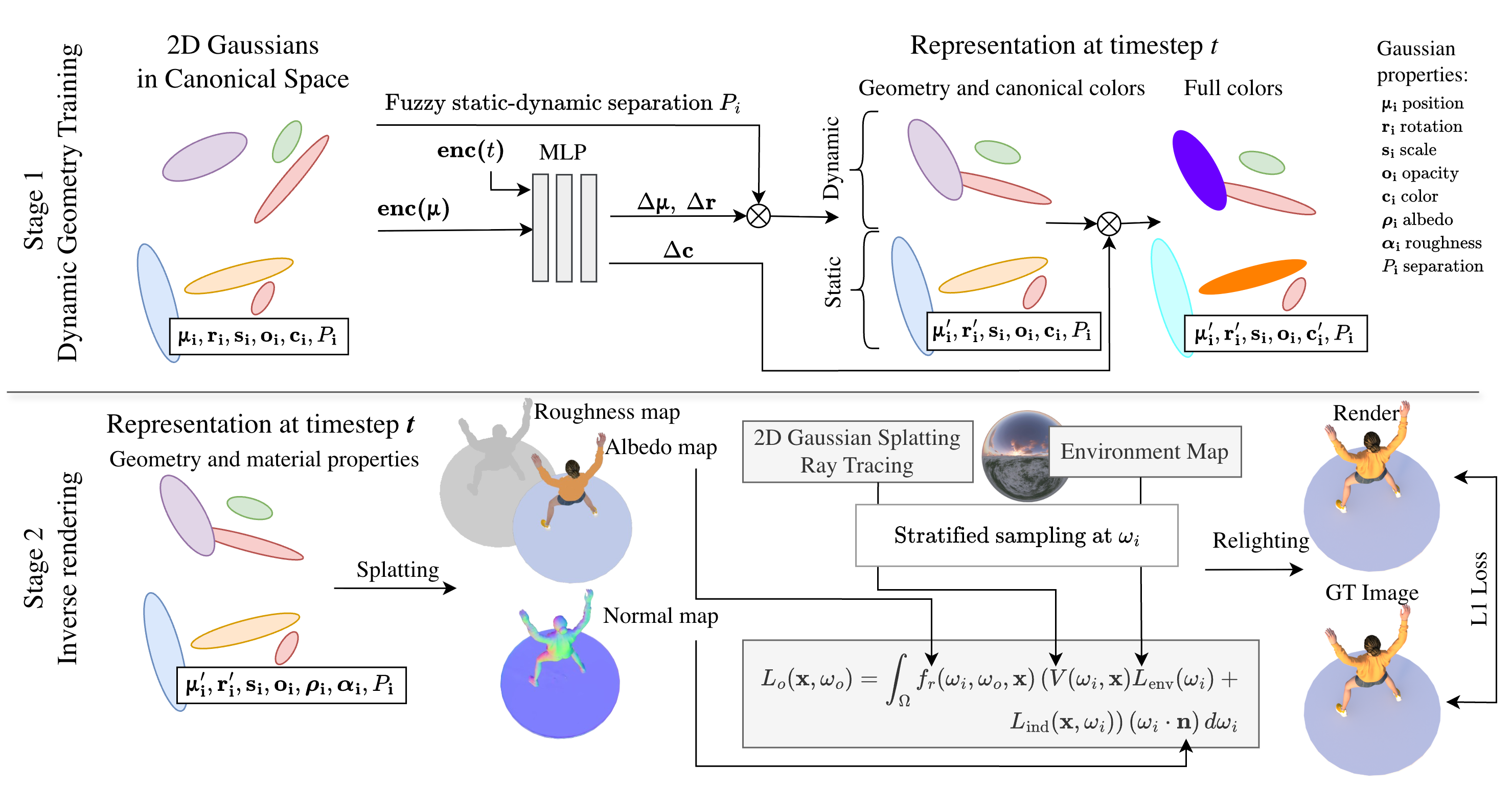}
  \vspace{-2em}
  \caption{\textbf{Method overview.} In Stage 1, a dynamic representation of 2D Gaussians, tailored for the relighting task, is trained. 
  The canonical color from Stage 1 serves as a starting point for albedo and is further optimized in Stage 2, together with roughness and the unknown environment map. In Stage 2, we rasterize the albedo, roughness and normal maps, then perform stratified sampling from the environment map \asia{$L_{\text{env}}$} and apply ray tracing to compute per-pixel visibility $V$ \asia{and indirect light $L_{\text{ind}}$}. L1 Loss is computed between the rendered pixels and the ground truth pixels.}
  \label{fig:method}
\vspace{-1.0em}
\end{figure*}

\paragraph{Inverse Rendering.}
Inverse rendering seeks to decompose a scene into its geometry, material properties, and lighting effects based on input images. A key challenge lies in the inherent ambiguity between observed photometric effects and the true material and lighting parameters, often resulting in multiple plausible solutions to the rendering equation. At the same time, modeling physical conditions is crucial for relighting optimized scenes. To this end, several methods have incorporated spatially-varying BRDF parameters into neural representations \cite{nerd, yao2022neilf, zhang2023neilf++, nerv, renerf}. For example, NERD leverages multiview supervision under varying illumination and employs a path-traced differentiable renderer. TensoIR~\cite{TensorIR} utilizes TensorRF~\cite{TensorRF} tri-plane representation for efficient computation of visibility and indirect lighting by ray-tracing. 

More recent methods explore inverse rendering using Gaussian Splatting, aiming to optimize material properties for each Gaussian. Some approaches focus solely on modeling reflective properties~\cite{ye2024gsdr_refl, yao2024refGS_refl, jiang2024gaussianshader}, while others utilize the full rendering equation, which requires accurate visibility estimation. In GS³~\cite{GS³}, occlusions are handled via shadow splatting, allowing for fast relighting but limited to OLAT-type training data. R3DG~\cite{R3DG2023} employs ray tracing for visibility estimation and baking, computing shading individually per Gaussian. GI-GS~\cite{GI-GS}, IRGS~\cite{irgs}, and GS-IR~\cite{liang2024gsirgs-ir} adopt a deferred shading approach: they first rasterize maps into a G-buffer, then apply the full rendering equation for shading. Notably, IRGS~\cite{irgs} leverages 2DGS with a differentiable ray tracer and addresses the computational overhead via stratified relighting in each iteration. Note that all of the above models focus on static scenes and thus do not leverage information from the scene motion. Other methods~\cite{URAvatar, hong2025beam, codec_avatars, WangARXIV2025_relightable_full_body, zhao2024surfel_human, zhan2024interactive_rel_human_avatars}, although designed to handle relightable dynamics, are constrained by human-pose priors and typically require either known training light conditions or access to large datasets with diverse lighting. In contrast to these approaches, our method learns lighting from a scene without the need to observe it under multiple illuminations and with no assumption on object category. \cref{tab:methods_datasets} summarizes the differences between the discussed methods and available datasets. %

\section{Method}
\ours~operates in a two-stage setup. In the first stage, we perform a 3D reconstruction of the scene. This process combines creation of Gaussians with learning a deformation network that models the scene's dynamics.
In the second stage, we take the geometry and deformation learned in the first stage, which are now frozen, and jointly optimize the illumination and material parameters of the scene. An overview of our method is presented in \cref{fig:method}.

\subsection{Preliminaries}
\label{sec:preliminaries}

\paragraph{2D Gaussian Splatting (2DGS).} 2DGS~\cite{2DG} is particularly well-suited for view synthesis and relighting, thanks to its ability to produce smooth and accurate surface normals. 2DGS represents a scene as a collection of flat 2D Gaussians embedded in 3D space. Each Gaussian is defined by a central point $\mathbf{\boldsymbol\mu} \in \mathbb{R}^3$, two tangential vectors $\mathbf{t}_u, \mathbf{t}_v \in \mathbb{R}^3$ which define the normal, and two scaling factors $s_u, s_v$ that control the spread in the local tangent plane. A 2D Gaussian is parameterized as follows:
\begin{equation}
G(u,v) = \mathbf{\boldsymbol\mu} + s_u \mathbf{t}_u u + s_v \mathbf{t}_v v.
\end{equation}
For rendering, each disk is projected to screen space via a homogeneous transformation. An explicit ray-splat intersection computes the local $(u,v)$ coordinates for each pixel, and the projected value is evaluated using a screen-space filter. Gaussians are alpha-composited front-to-back based on depth ordering to accumulate color per pixel. For more implementation details, we refer the reader to~\cite{2DG}.

\paragraph{Rendering Equation.} In Stage 2 of our pipeline, we relight the reconstructed scene under a given illumination. In computer graphics, the appearance of the surface is governed by the classical rendering equation~\cite{Kajiya_rendering_equation}, which models the interaction between the properties of light and material:
\begin{equation}
\label{eq:rendering_eq}
\begin{split}
L_o(\mathbf{x},\omega_o)
&= \int_{\Omega} f_r(\omega_i,\omega_o, \mathbf{x})
\big(V(\omega_i,\mathbf{x})L_{\text{env}}(\omega_i) \\
& \qquad + L_{\text{ind}}(\mathbf{x},\omega_i)\big)
(\omega_i\!\cdot\!\mathbf{n})\, \mathrm{d}\omega_i,
\end{split}
\end{equation}
where $L_o(\mathbf{x}, \omega_o)$ denotes the outgoing radiance at point $\mathbf{x}$ in direction $\omega_o$, 
$f_r(\omega_i, \omega_o, \mathbf{x})$ represents the bidirectional reflectance distribution function (BRDF), 
$L_{\text{env}}(\omega_i)$ is the incoming environment radiance from direction $\omega_i$, 
$\mathbf{n}$ is the surface normal at $\mathbf{x}$, $L_{\text{ind}}(\mathbf{x}, \omega_i)$ is the indirect illumination term, $V(\omega_i, \mathbf{x})$ is the visibility of environment light from the point $\mathbf{x}$ in the direction $\omega_i$, and $\Omega$ denotes the hemisphere oriented around $\mathbf{n}$. The BRDF describes the amount of light reflected from direction $\omega_i$ towards $\omega_o$ for the material at position $\mathbf{x}$.

Most inverse rendering tasks aim to decompose and reconstruct scene components by estimating material properties ($f_r$) and environment illumination ($L_{env}$) from observed images. Following the simplified Disney BRDF model \cite{disney_burley2012physically}, we parametrize bidirectional reflectance distribution function with albedo $\mathbf{\rho}$ and roughness $\alpha$. While those values change throughout the scene and thus depend on the position $\mathbf{x}$ we omit the position here for brevity. The final BRDF which combines diffuse and specular terms is:
\begin{equation}
    f_r(\omega_i, \omega_o){=}\frac{\mathbf{\rho}}{\pi} + \frac{D(\mathbf{\rho};\alpha)F(\omega_o, \mathbf{h};\mathbf{\rho})G(\omega_i, \omega_o, \mathbf{h};\mathbf{\alpha})}{4 (\omega_i \cdot \mathbf{n})( \omega_o \cdot \mathbf{n})},
\end{equation}
where $D$, $F$, and $G$ denote the normal distribution function, the Fresnel term, and the geometry term, respectively, and $\mathbf{h} = \frac{\omega_i + \omega_o}{\|\omega_i + \omega_o\|}$, for details see~\cite{GI-GS}. We further assume that elements of the scene do not emit light.

\begin{figure*}
\centering
\includegraphics[width=0.87\linewidth]{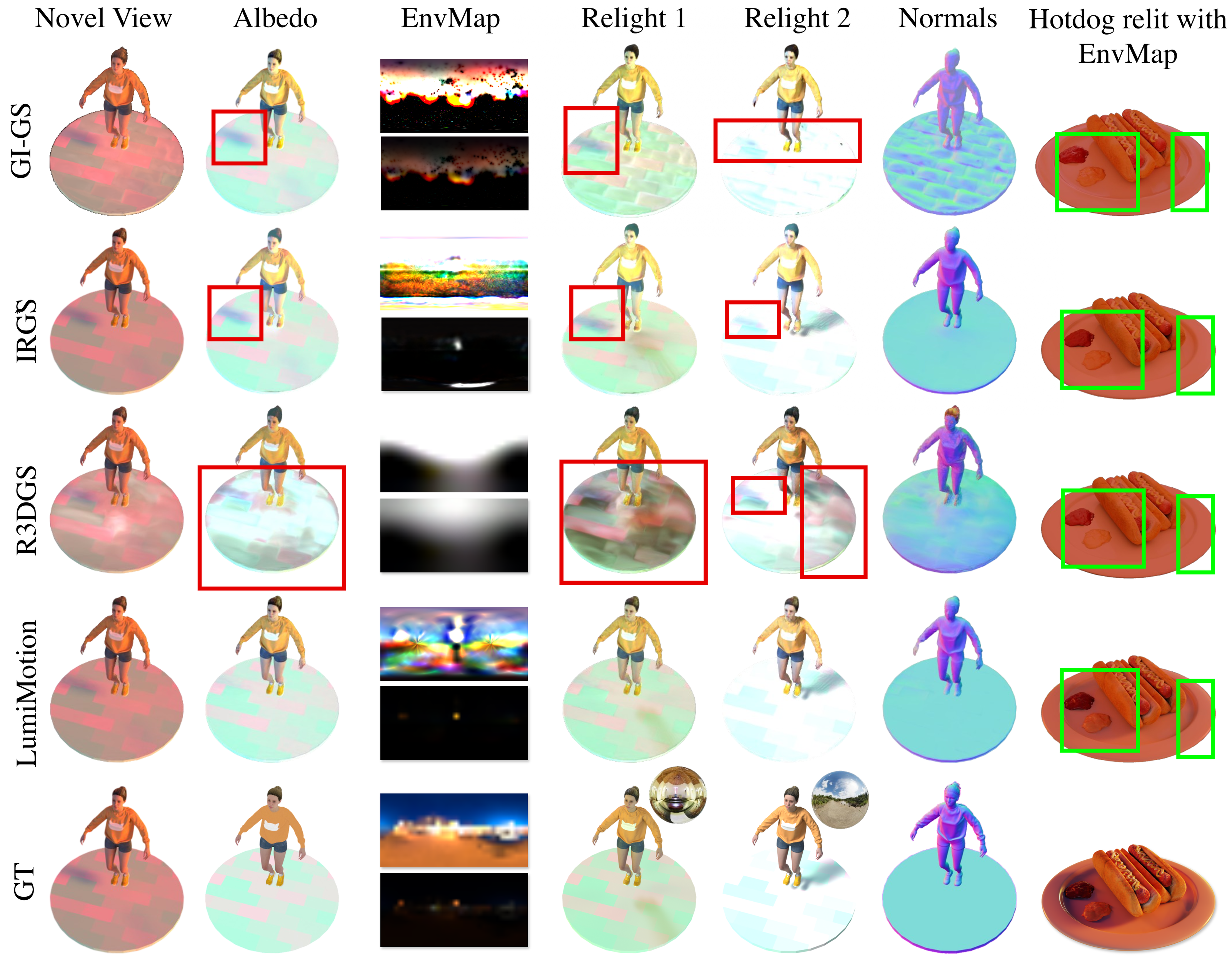}
\caption{\textbf{Qualitative comparison of novel view synthesis, relighting, components and estimated lighting.} \asia{All renders are shown from a novel viewpoint.} Our method not only achieves stronger relighting capabilities and effectively removes shadows from albedo, but also estimates lighting more accurately. The presented environment map (top shows the map truncated to $[0, 1]$ while bottom shows it scaled so that the maximum value is $1$) captures the main light direction more accurately and relights new item more realistically than environment maps estimated from static baselines. \textbf{Please zoom in for details.}}
\label{fig:relight_components_example}
\vspace{-1.0em}
\end{figure*}

\subsection{Stage 1: Dynamic Geometry Learning for Relighting} \label{sec:stageonemethod}

We base our method on 2D Gaussian Splatting (2DGS), which provides accurate surface normals that are critical for separating illumination from materials. To capture dynamic scene changes over time, following classical dynamic scene modeling approaches~\cite{Yang_2024_CVPR_deformable_gs, Wu_2024_CVPR_4dgaussians}, we use a multilayer perceptron (MLP) to predict Gaussian transformations.

Given a timestep $t \in [0, 1]$ and the canonical Gaussian position $\mathbf{\boldsymbol\mu} = (x, y, z)$, the MLP predicts changes in position $\Delta \mathbf{\boldsymbol\mu}\in \mathbb{R}^3$, rotation $\Delta \mathbf{r}\in \mathbb{R}^3$, and color $\Delta \mathbf{c}\in \mathbb{R}^3$:
\begin{equation}
(\Delta \mathbf{\boldsymbol\mu}, \Delta \mathbf{r}, \Delta \mathbf{c}) = \text{MLP}(\text{enc}(t), \text{enc}(\mathbf{\boldsymbol\mu})),
\end{equation}
where $\text{enc}(\cdot)$ denotes positional encoding \cite{nerf}. The timestep $t$ denotes a moment in time for the dynamic scene being modeled. Note that we choose not to model changes of opacity or scale of the Gaussians as this would allow the MLP to make objects appear and disappear instead of moving them through space. This, in turn, would go against our main goal, which is to recover illumination and materials with the use of motion in the scene.

\paragraph{Static-dynamic fuzzy separation.} To compute the actual Gaussian position and rotation at time $t$, we could simply add the predicted deltas to the canonical values. However, we notice that accurately modeling element dynamics is crucial for correct albedo estimation and can be flawed in textureless areas. For example, moving shadows on a flat surface can be explained either by color changes or by moving/disappearing Gaussians, or a combination of both. Importantly, a moving Gaussian representing a moving shadow cannot be assigned a stable albedo color in Stage 2. Thus, we aim to separate static and dynamic components explicitly in Stage 1.

To achieve such separation, we introduce an auxiliary per-Gaussian variable $P$ that indicates whether the Gaussian belongs to the static or dynamic group. We sample $P$ using a Binary Concrete distribution~\cite{binary_concrete}, a continuous relaxation of a Bernoulli distribution that concentrates most mass near 0 or 1. The relaxed variable $\tilde{P}$ is defined as:
\begin{align}
\tilde{P} &= \text{sigmoid}\left( \frac{1}{T} \left( \log(|P|) + \log(U) - \log(1 - U) \right) \right), \nonumber \\
U &\sim \text{Uniform}(0, 1).
\end{align}
where $T$ is the temperature hyperparameter. We set $T=0.5$, encouraging a \textbf{near-binary separation}. During inference, we fix $U = 0.5$ to deterministically obtain $\tilde{P}$.

The final Gaussian position $\mathbf{\boldsymbol\mu}'$ and rotation $\mathbf{r}'$ at time $t$ are then computed as:
\begin{equation}
\mathbf{\boldsymbol\mu}' = \mathbf{\boldsymbol\mu} + \tilde{P} \Delta \mathbf{\boldsymbol\mu}, \quad \mathbf{r}' = \mathbf{r} + \tilde{P} \Delta \mathbf{r}.
\end{equation}
This formulation along with the regularization on $P$ explained below ensures that dynamic changes are applied selectively, leading to more accurate modeling of scene geometry in time, which is essential for precise relighting and material decomposition in the second stage.

\paragraph{Modeling Temporal Color Variation.} Since we focus on dynamic scenes under static lighting, we allow Gaussian colors to vary over time. We observe that significant color changes typically arise from two sources: (i) moving shadows affecting both static and dynamic elements, and (ii) changes in incident illumination on dynamic elements due to motion. To model these effects, we define the color at time $t$ as:
\begin{equation}
\mathbf{c}' = \mathbf{c} (1 - \Delta \mathbf{c}),
\end{equation}
where $\mathbf{c}$ is Gaussian canonical color. 
We use multiplicative change to mimic how light affects surfaces (see~\cref{eq:rendering_eq}), while $(1 - \Delta \mathbf{c})$ allows for applying regularization on excessive color variation. 
Such formulation captures effects like moving shadows and illumination changes on dynamic elements.
The canonical color $\mathbf{c}$ approximates a pseudo-albedo that serves as the initial estimate for material decomposition in Stage~2.

\begin{figure*}
\centering
\includegraphics[width=\linewidth]{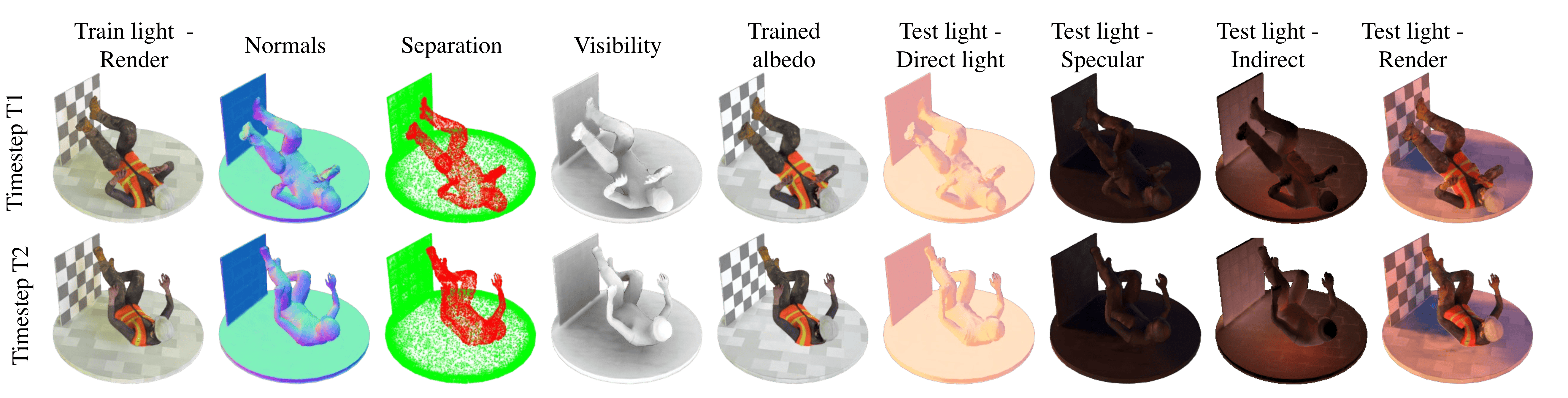}
\caption{\textbf{Qualitative evaluation of components for a dynamic scene for different timesteps.} \ours{} maintains consistent normals across timesteps, accurately separates static elements, and produces temporally consistent relighting.}
\label{fig:dynamic}
\vspace{-1.0em}
\end{figure*}

\paragraph{Training Loss.}
Following 2DGS \cite{2DG}, we apply the reconstruction loss $\mathcal{L}_c$, along with normal consistency loss $\mathcal{L}_n$, which aligns the rendered normal map with the underlying surface geometry, and depth distortion loss $\mathcal{L}_d$, which encourages tight spatial concentration of the Gaussians. 

To handle floating Gaussians, we apply a binary cross-entropy loss $\mathcal{L}_o$ with respect to the foreground mask $\mathcal{M}$:
\begin{equation}
\mathcal{L}_o = -\mathcal{M} \log \mathcal{O} - (1 - \mathcal{M}) \log (1 - \mathcal{O}),
\end{equation}
where $\mathcal{O}$ is per-pixel accumulated gaussian opacity.
In addition to the above, we introduce losses specific to dynamic modeling.  

\paragraph{Static-Dynamic Separation Loss $\mathcal{L}_P$.} 
To encourage Gaussians to remain static whenever possible, we apply an $L1$ penalty on the dynamic assignment variable $P$:
\begin{equation}
\mathcal{L}_P = \frac{1}{N} \sum_{i=1}^{N} |P_i|,
\end{equation}
where $P_i$ is the predicted probability of Gaussian $i$ is dynamic, and $N$ is the number of Gaussians.
Minimizing $\mathcal{L}_P$ directly encourages $P_i$ to be close to $0$, favoring static representations and reducing unnecessary dynamic modeling.

\paragraph{Color and Position Change Regularization.} To additionally discourage the model from predicting unnecessary position and color changes, we apply \(L2\) regularization on both the predicted color and position deltas, defined as
\begin{equation}
\mathcal{L}_{\Delta c} = \frac{1}{N} \sum_{i=1}^{N} \|\Delta \mathbf{c}_i\|^2, \quad
\mathcal{L}_{\Delta \mu} = \frac{1}{N} \sum_{i=1}^{N} \|\Delta \mathbf{\boldsymbol\mu}_i\|^2,
\end{equation}
penalizing excessive color variation and spatial movement of Gaussians.

\begin{table*}
\centering
\caption{\textbf{Quantitative results for albedo estimation and relighting.} Static methods use only one timestep with multiple views of the scene. \ours{} uses a dynamic scene for training while testing is done on the same views and the same single timestep as in the static setting. Results are grouped by train-test light setting. Although training on dynamic scenes is \textbf{inherently more challenging}, \ours{} significantly outperforms albedo  estimation tasks on all metrics across all baselines. \marek{Notably, for both tasks, we achieve a substantial improvement in LPIPS which is a metric that corresponds well to perceptual quality.}}
\resizebox{0.99\linewidth}{!}{

\begin{tabular}{clcccccc} %
\toprule
\multirow{2}[2]{*}{Env Lights} &\multirow{2}[2]{*}{Method}
& \multicolumn{3}{c}{Albedo} 
& \multicolumn{3}{c}{Relight} 
\\
\cmidrule(lr){3-5}\cmidrule(lr){6-8}
& & PSNR $\uparrow$ & SSIM $\uparrow$ & LPIPS $\downarrow$
& PSNR $\uparrow$ & SSIM $\uparrow$ & LPIPS $\downarrow$ 
\\
\midrule
\textbf{Dam Wall} &R-3DGS  & 20.744 {$\pm$ 0.661} & 0.900 {$\pm$ 0.013} & 0.128 {$\pm$ 0.031}& 21.220 {$\pm$ 1.843} & 0.915 {$\pm$ 0.016}& 0.112 {$\pm$ 0.028} %
\\
\textbf{$\downarrow$} &GI-GS   & 20.943 {$\pm$ 1.747} & 0.906 {$\pm$ 0.014} & 0.105 {$\pm$ 0.023} & 18.434 {$\pm$ 1.681} & 0.868 {$\pm$ 0.023} & 0.139 {$\pm$ 0.032} %
\\
\textbf{Harbour Sunset} &IR-GS   & \cellcolor{red!15}{22.888} {$\pm$ 1.559} & \cellcolor{red!15}{0.936} {$\pm$ 0.013}  & \cellcolor{red!15}{0.076} {$\pm$ 0.023} & \cellcolor{red!40}{26.177} {$\pm$ 1.606} & \cellcolor{red!40}{0.953} {$\pm$ 0.011}  & \cellcolor{red!15}{0.064} {$\pm$ 0.018} %
\\
&\ours{} & \cellcolor{red!40}{27.268} {$\pm$ 1.568}  & \cellcolor{red!40}{0.952} {$\pm$ 0.007} & \cellcolor{red!40}{0.069} {$\pm$ 0.024} & \cellcolor{red!15}{26.037} {$\pm$ 0.579} & \cellcolor{red!15}{0.928} {$\pm$ 0.007}& \cellcolor{red!40}{0.060} {$\pm$ 0.012} %
\\
\midrule
\textbf{Chapel Day} &R-3DGS  & 22.463 {$\pm$ 2.001} & 0.927 {$\pm$ 0.017} & 0.096 {$\pm$ 0.038}& 22.282 {$\pm$ 2.806} & \cellcolor{red!15}{0.943} {$\pm$ 0.012} & 0.081 {$\pm$ 0.030} %
\\
\textbf{$\mathbf{\downarrow}$} &GI-GS   & \cellcolor{red!15}{24.733} {$\pm$ 2.862} & 0.955 {$\pm$ 0.014} & 0.056 {$\pm$ 0.016} & 22.673 {$\pm$ 1.513} & 0.880 {$\pm$ 0.015} & 0.125 {$\pm$ 0.022} %
\\
\textbf{Golden Bay} &IR-GS   & 23.769 {$\pm$ 2.732} & \cellcolor{red!15}{0.956} {$\pm$ 0.015} & \cellcolor{red!15}{0.053} {$\pm$ 0.017} & \cellcolor{red!15}{28.157} {$\pm$ 1.978} & \cellcolor{red!40}{0.966} {$\pm$ 0.009} & \cellcolor{red!15}{0.046} {$\pm$ 0.020} %
\\
&\ours{} & \cellcolor{red!40}{30.838} {$\pm$ 1.798} & \cellcolor{red!40}{0.973} {$\pm$ 0.007}  & \cellcolor{red!40}{0.036}{$\pm$ 0.014}  & \cellcolor{red!40}{28.563} {$\pm$ 0.478} & 0.939 {$\pm$ 0.011}& \cellcolor{red!40}{0.041} {$\pm$ 0.007} %
\\
\midrule
\textbf{Golden Bay} &R-3DGS  & 19.945 {$\pm$ 1.124} & 0.899{$\pm$ 0.018} & 0.133 {$\pm$ 0.041} & 19.563 {$\pm$ 1.874}& 0.918 {$\pm$ 0.013} & 0.118 {$\pm$ 0.033} %
\\
\textbf{$\mathbf{\downarrow}$}&GI-GS   & \cellcolor{red!15}{21.295} {$\pm$ 2.930} & 0.932 {$\pm$ 0.020} & 0.087 {$\pm$ 0.025} & 17.636 {$\pm$ 2.293} & 0.823 {$\pm$ 0.029} & 0.132 {$\pm$ 0.030} %
\\
\textbf{Dam Wall} &IR-GS   & 20.910 {$\pm$ 1.379} & \cellcolor{red!15}{0.937} {$\pm$ 0.013} & \cellcolor{red!15}{0.082} {$\pm$ 0.024} & \cellcolor{red!15}{25.009} {$\pm$ 1.615} & \cellcolor{red!40}{0.955} {$\pm$ 0.011} & \cellcolor{red!15}{0.060} {$\pm$ 0.014} %
\\
&\ours{} & \cellcolor{red!40}{27.929} {$\pm$ 1.932}  & \cellcolor{red!40}{0.959}{$\pm$ 0.010} & \cellcolor{red!40}{0.058} {$\pm$ 0.019} & \cellcolor{red!40}{25.405} {$\pm$ 0.690} & \cellcolor{red!15}{0.936} {$\pm$ 0.013}& \cellcolor{red!40}{0.048} {$\pm$ 0.009} %
\\
\bottomrule
\end{tabular}
}
\vspace{-0.1cm}
\label{tab:quantitative_eval}
\end{table*}

\paragraph{Overall Loss.}
The total loss for the first stage training is then:
\begin{equation}
\begin{split}
\mathcal{L}^1 = \mathcal{L}_c + \lambda_n \mathcal{L}_n + \lambda_d \mathcal{L}_d + \lambda_o \mathcal{L}_o + \lambda_P \mathcal{L}_P\\
+ \lambda_{\Delta c} \mathcal{L}_{\Delta c} + \lambda_{\Delta \mu} \mathcal{L}_{\Delta \mu},
\end{split}
\end{equation}
where $\lambda_n, \lambda_d, \lambda_o, \lambda_P, \lambda_{\Delta c}, \lambda_{\Delta \mu}$ are weighting coefficients. Please see supplementary materials for their weights.

\subsection{Stage 2: Inverse Rendering}

In the second stage, we perform inverse rendering to decompose the scene into material properties and environment lighting. To model materials, each 2D Gaussian is assigned a diffuse albedo $\mathbf{\rho}$ that is initialized with the canonical color $\mathbf{c}$ from the first stage, and roughness $\alpha$. These properties remain constant for each timestep $t$. In Stage 2, color changes arise solely from the rendering equation, which determines light–surface interaction and thus the $\Delta\mathbf{c}$ output of the MLP is not used. Environment lighting $L_{env}$ is modeled using an image where each pixel corresponds to light intensity and color from a direction $\omega_i$. During Stage 2 we jointly optimize $\mathbf{\rho}$ and $\alpha$ for each Gaussian as well as a single $L_{env}$ for the whole scene. Further details about optimized parameter set and gradient flow are available in supplementary.

\begin{figure}
\centering
\includegraphics[width=\linewidth]{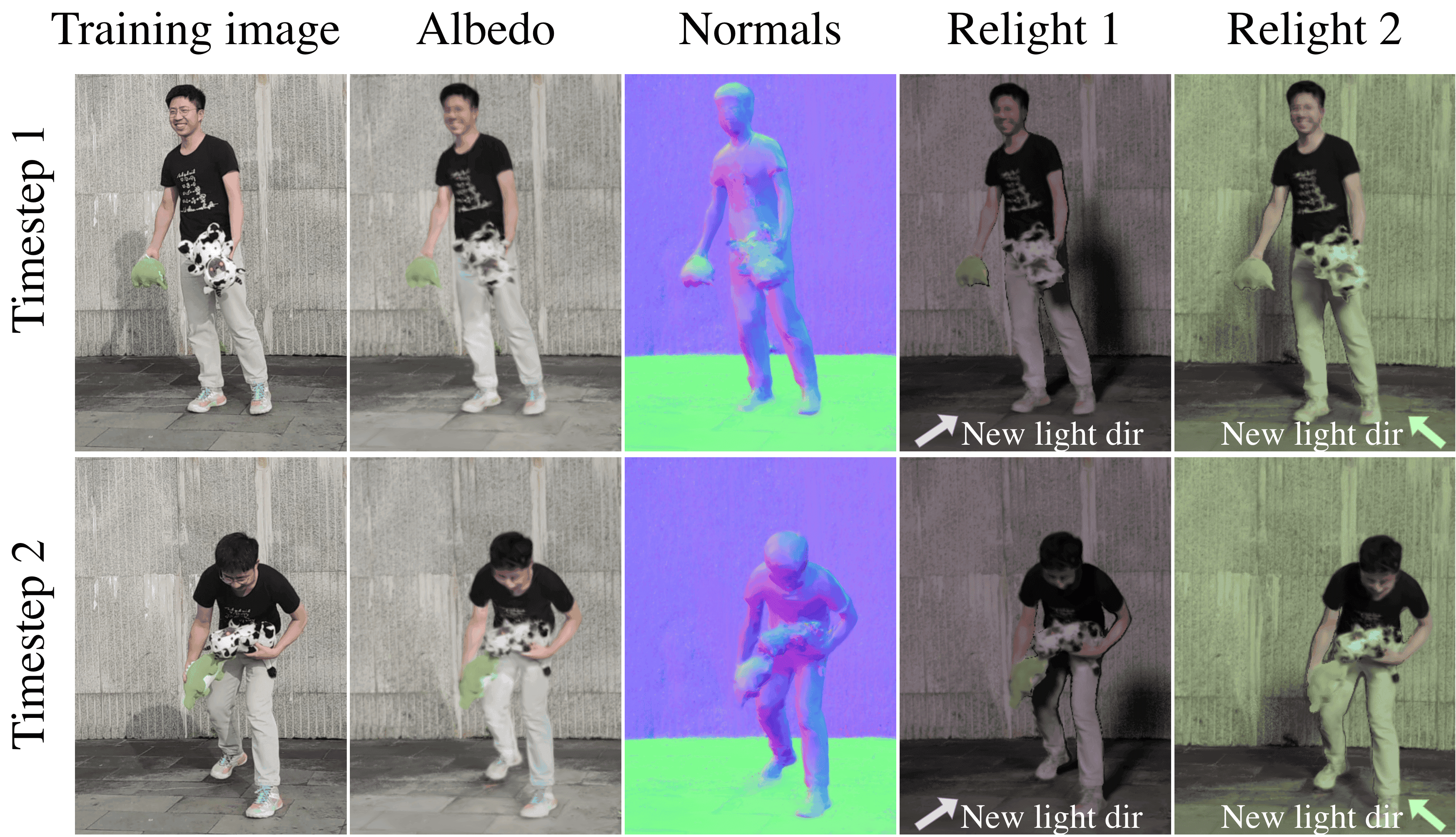}
\caption{\textbf{\ours{} on real-world ENeRF scene.} The sharp shadow cast by the dynamic actor is largely removed from albedo. Under novel lighting, the actor appears realistic, casting shadows in appropriate locations.}
\label{fig:separation_weightnew shadows in appropriate locations.}
\label{fig:real_data2}
\vspace{-1.0em}
\end{figure}

When rendering the scene, similarly to \cite{irgs, GI-GS}, we apply the rendering equation after rasterization rather than per-Gaussian. This approach allows shading effects such as shadows to appear at each pixel, rather than being limited to per-Gaussian granularity. To obtain per-pixel material values we alpha-blend the albedo and roughness attributes across Gaussians during rasterization.

The incident radiance $L_i$ at surface point $\mathbf{x}$ along direction $\omega_i$ is represented by the sum $V(\omega_i, \mathbf{x}) L_{\text{env}}(\omega_i) + L_{\text{ind}}(\omega_i, \mathbf{x})$, where %
the visibility term $V(\omega_i, \mathbf{x}) \in \{0, 1\}$ is obtained via 2D Gaussian ray tracing from $\mathbf{x}$ in direction $\omega_i$. A low value of $V(\omega_i, \mathbf{x})$ indicates that the light from $L_{env}(\omega_i)$ is occluded before reaching point $\mathbf{x}$.
\asia{We compute indirect term $L_{\text{ind}}$ similarly to \cite{irgs} where indirect light values are traced: during training, RGB values used in ray tracing correspond to the colors $\mathbf{c}$ from the first stage, while
for inference under novel lighting, we use colors evaluated from the rendering equation.
} 
We employ Monte Carlo integration with uniform stratified sampling selecting $N_r$ ray directions over the hemisphere to efficiently evaluate the rendering equation. The final RGB output of the rendering function can be written as: 
\begin{align}
\textbf{c}_{\text{pbr}}(\omega_o)
&= \frac{2\pi}{N_r}\sum_{i=1}^{N_r}
   f_r(\omega_i,\omega_o,\mathbf{x}) L_{i}(\omega_i,\mathbf{x}) 
   (\omega_i \cdot \mathbf{n})
\end{align}
where $L_i$ denotes incident radiance. Following \cite{irgs}, we also randomly sample N pixels per iteration to reduce computation time.  

Stage 2 combines three losses: (1) ${L_c}$ loss from Stage 1, (2) ${L1}$ loss for Stage 2 renders against GT pixels, (3) ${L2}$ regularization with small weight ${\lambda_{env}}$ that penalizes high values in the lower region of $L_{env}$. ${L_c}$ is computed between GT images and pure Gaussian splatting renders and it constrains the fine-tuned Gaussian parameters, preventing them from deviating excessively from their Stage 1 values. ${L1}$ is the only loss used to supervise the $\mathbf{c_{{\text{pbr}}}}$.

\section{Experiments and Results}
\vspace{-0.5em}
\paragraph{Datasets.} To thoroughly evaluate our method under controlled relighting and shadow-casting conditions, we introduce a novel synthetic benchmark consisting of \textbf{20 variations}: five distinct scenes, each rendered under four different lighting environments (`Harbour Sunset', `Dam Wall', `Golden Bay', and `Chapel Day'). The scenes differ in object types, motion patterns, and levels of surface specularity. For every scene, both static and dynamic versions are provided to enable fair comparison across temporal settings. For dynamic version we have D-NeRF~\cite{pumarola2021d}-like captures: one view per one timestep. The static dataset captures a single timestep from the same camera poses as the dynamic version. This setup enables training both static and dynamic models on comparable data, and allows direct evaluation using identical camera views and time steps. Test set consists of novel views. Additional details about the dataset are provided in the \textbf{Supplementary Material}.

To qualitatively assess performance in real-world conditions, we use two scenes from ENeRF dataset~\cite{enerf}, an outdoor dataset which captures dynamic people casting prominent shadows from 18 cameras. Its multi-view nature supports training and evaluation of static and dynamic models.

\paragraph{Experimental setup.} For synthetic data, we evaluate key aspects of our method: albedo estimation and relighting quality. We report PSNR, SSIM \cite{ssim}, and LPIPS \cite{zhang2018perceptual_lpips} as evaluation metrics. For each scene in our dataset, we conduct experiments under three configurations, selecting one light for training and another for testing. Detailed training parameters and extended evaluation including videos are provided in the Supplementary Material. All experiments are conducted on an NVIDIA RTX 3090, with training for both stages taking approximately 1.2 hours per synthetic scene.

\begin{figure}
\centering
\includegraphics[width=\linewidth]{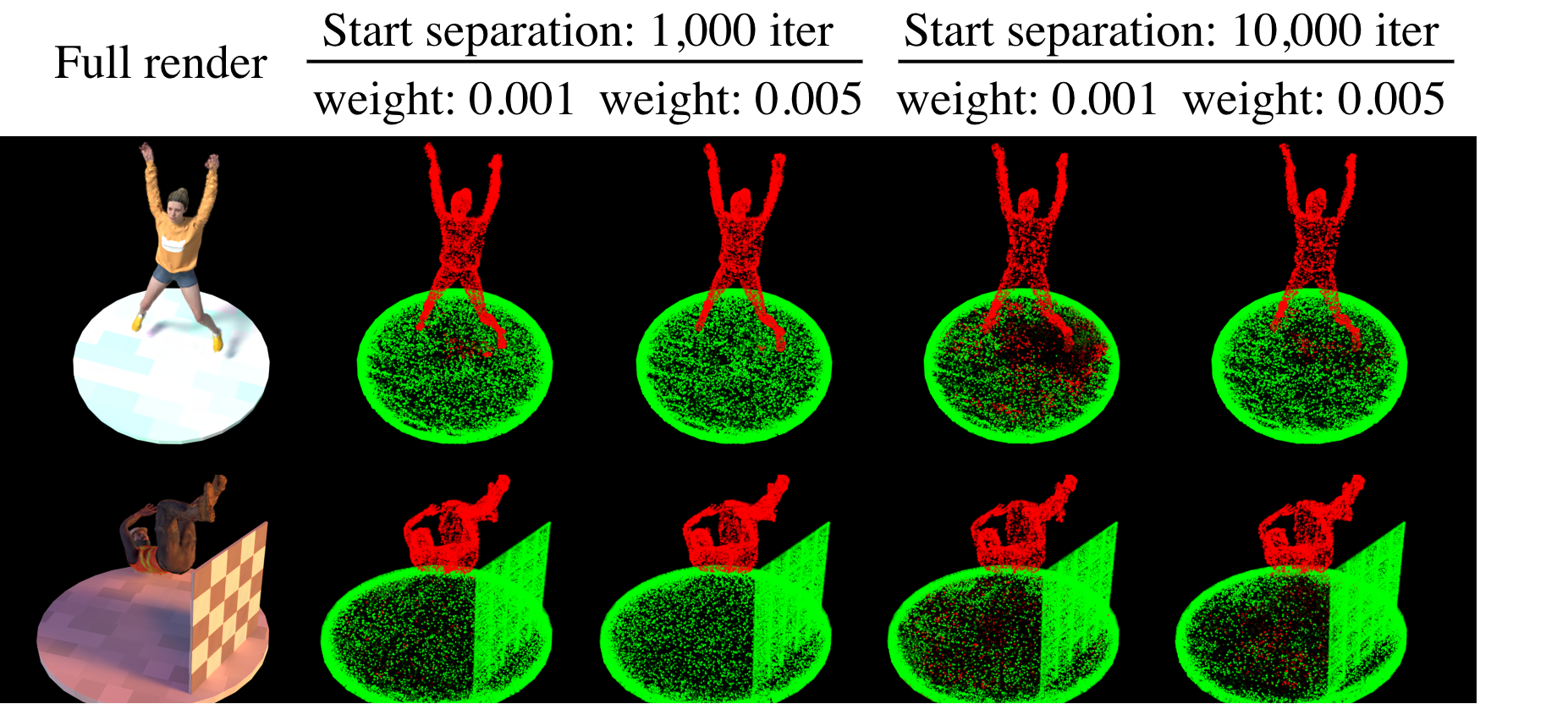}
\caption{\textbf{Influence of separation loss parameters.} The dynamic objects in the scene remain correctly classified as dynamic across tested separation weights and initialization times. However, shadows on static parts of the scene (e.g. ground) can be misclassified as dynamic Gaussians (here mostly visible for start 10,000 iter). Increasing the separation weight or starting the separation earlier helps to reduce such misclassifications. \textbf{Zoom in for details.}}
\label{fig:separation_weight}
\vspace{-0.1cm}
\end{figure}

\begin{figure}
\centering
\includegraphics[width=\linewidth]{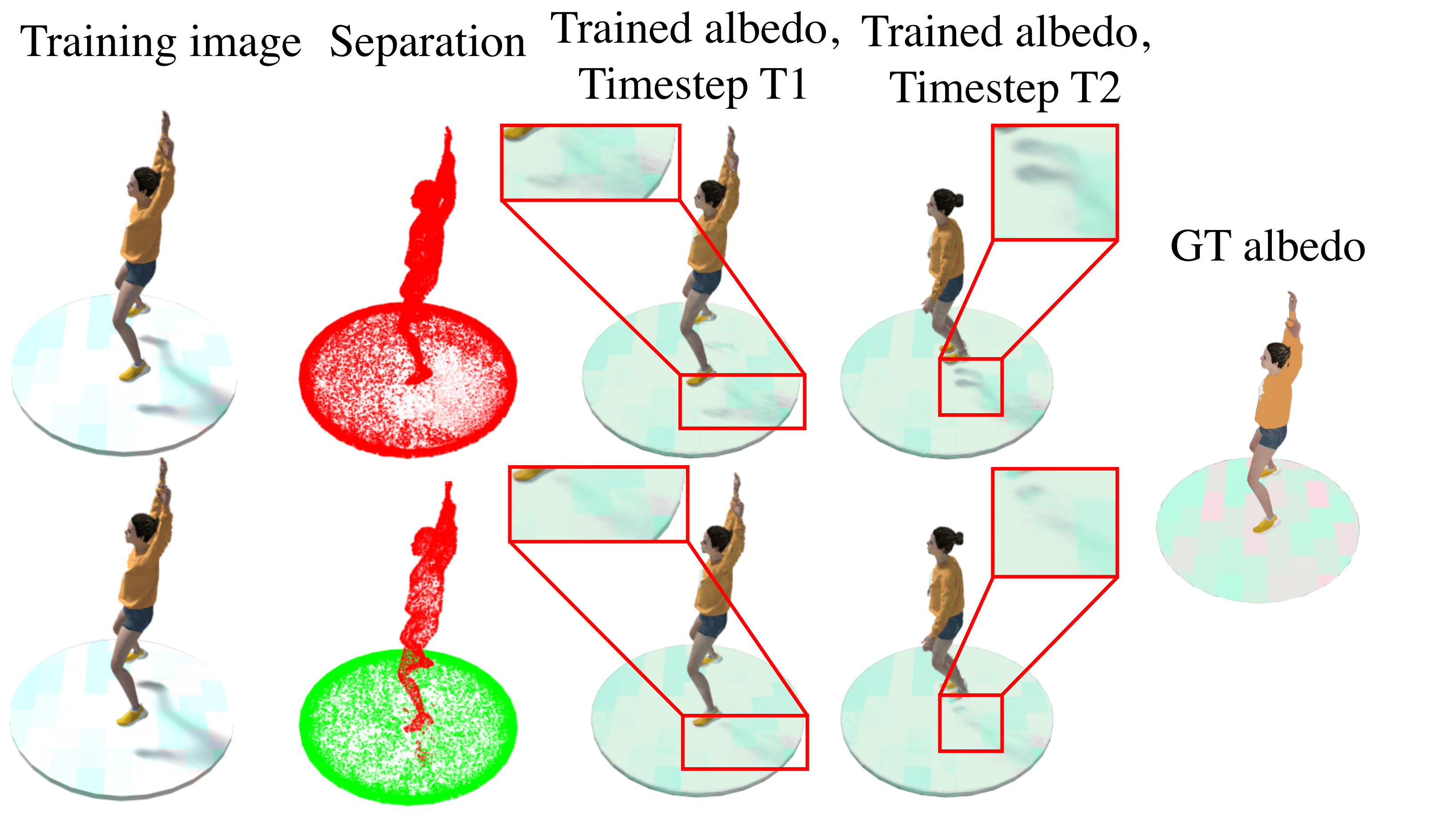}
\caption{\textbf{Ablation study of static-dynamic separation.} The demonstrated scene is characterized with strong directional light and strong shadows. Without the static-dynamic separation (top), the Gaussians reconstruct a strong shadow that follows the object motion, which degrades albedo optimization \asia{in the second stage}. When separation is enabled, the shadows are modeled by $\Delta \mathbf{c}$ instead of $\Delta\mathbf{\boldsymbol\mu}$ which enables correct albedo optimization. \textbf{Please zoom in for details.}}
\label{fig:stat_dyn_ablation}
\vspace{-0.3cm}
\end{figure}

\paragraph{Results.} Quantitative comparisons against state-of-the-art methods are summarized in~\cref{tab:quantitative_eval}, demonstrating our method's strong performance across the evaluation metrics. We achieve strong performance in albedo estimation, surpassing \textbf{all baselines across all metrics} by a large margin, which demonstrates the enhanced capability of our method in removing light-related artifacts. For relighting, our metrics surpass the baselines in most cases. With regard to LPIPS---the metric that best corresponds to perceptual quality---\ours~performs best in all cases, with an average improvement of $15\%$. Relighting task is inherently more difficult for dynamic scenes - it depends strongly on properly estimated normals, which are more challenging to obtain due to the added constraint of temporal consistency. It is notable that our approach performs competitively on the relighting task, highlighting its capability even under the complexities of a dynamic setup.
We present qualitative results in~\cref{fig:relight_components_example} and highlight improvements in relighting fidelity, albedo consistency and environment map reconstruction. Please note that, thanks to dynamics, we can better estimate the direction of incoming light, and also prevent the model from baking shadows in the base color.  \textbf{We provide extended results in the Supplementary Material.}

We evaluate our approach on two clips from the \textbf{challenging real-world} ENeRF dataset, where sharp shadows are cast by moving actors. This setup enables us to assess our method’s robustness in handling real-world lighting variations and dynamic geometry. Since it is a multiview setup, we can compare our method with IRGS which is the most recent baseline. See~\cref{fig:real_world_qual} where, relative to IRGS, we remove the majority of shadows from the albedo from the first scene. Additionally, unlike IRGS our method does not exhibit artifacts when rendering with specular component. The second scene is presented in \cref{fig:real_data2}.

In \cref{fig:dynamic}, we demonstrate that our method produces coherent renderings across timesteps, with smooth normals, consistent separation of dynamic elements, and shadows estimated in accordance with the moving geometry.

Finally, we perform an ablation study on key components of our pipeline (\cref{tab:ablation}). \cref{fig:separation_weight} illustrates the influence of separation parameters, showing how varying the separation weight and initialization affects the identification of dynamic elements. We show that Gaussians that move to simulate lighting effects, such as shadows, negatively impact albedo optimization. \cref{fig:stat_dyn_ablation} highlights the benefits of our separation strategy in addressing this issue.

\begin{table}
\centering
\caption{\textbf{Ablation study on Golden Bay → Dam Wall configuration,} characterized by strong directional train light. The additional components improve albedo and relight reconstruction quality.}
\vspace{-0.5em}
\resizebox{\linewidth}{!}{
\begin{tabular}{lcccccc}
\toprule
\multirow{2}[2]{*}{Method} & \multicolumn{3}{c}{\textbf{Albedo}} & \multicolumn{3}{c}{\textbf{Relight}} \\
\cmidrule(lr){2-4}\cmidrule(lr){5-7}
& PSNR $\uparrow$ & SSIM $\uparrow$ & LPIPS $\downarrow$ & PSNR $\uparrow$ & SSIM $\uparrow$ & LPIPS $\downarrow$ \\
\midrule
 w/o Stage 2 & 23.935 & 0.943 & 0.072 & 23.154 & 0.926 & 0.060\\
w/o $\Delta \mathbf{c}$ & 23.983 & 0.937 & 0.083 & 23.142 & 0.926 & 0.061 \\
w/o $\mathit{P}$ & 26.768 & 0.954 & 0.064 & 25.079 & 0.933 & 0.052 \\
full model & \textbf{27.929} & \textbf{0.959} & \textbf{0.058} & \textbf{25.405} & \textbf{0.936} & \textbf{0.048} \\
\bottomrule
\end{tabular}
}%
\label{tab:ablation}
\vspace{-1em}
\end{table}

\section{Conclusions and Limitations}
We proposed a two-stage inverse rendering framework that leverages dynamics as a supervisory signal to separate illumination from material properties. We further introduced a benchmark with scenes under various lighting and motion conditions, enabling systematic evaluation of our approach. \ours{} achieves improved relighting performance and more accurate material estimation compared to static-only baselines, proving that incorporating dynamic content is beneficial for these tasks. \textbf{Limitations.} We observe that accurate normal estimation and the quality of the learned deformations are critical to performance. In our framework, reconstruction quality remains limited, as temporally consistent and physically accurate motion and normal estimation in complex dynamic scenes is still an open challenge. Our simple separation strategy can generate artifacts when handling intricate dynamics, and more accurate supervision, such as optical flow, may be crucial. Additionally, the framework is sensitive to inaccurate camera pose estimation, sparse camera setups or inaccurate initialization.

\clearpage
\paragraph{Acknowledgements.} This paper received funding from the European Union’s Horizon 2020 research and innovation programme under grant agreement No 857533. The research is supported by Sano project carried out within the International Research Agendas programme of the Foundation for Polish Science, co-financed by the European Union under the European Regional Development Fund. The research was created within the project of the Minister of Science and Higher Education ”Support for the activity of Centers of Excellence established in Poland under Horizon 2020” on the basis of the contract number MEiN/2023/DIR/3796. The work of J. Kaleta was supported by National
Science Centre, Poland (grant no. 2022/47/O/ST6/01407). The work of K. Marzol was supported by the project \textit{Effective Rendering of 3D Objects Using Gaussian Splatting in an Augmented Reality Environment} (FENG.02.02-IP.05-0114/23), carried out under the First Team programme of the Foundation for Polish Science and co-financed by the European Union through the European Funds for Smart Economy 2021–2027 (FENG). The work of P. Wójcik was supported by the German Research Foundations (DFG) funded Collaborative Research Center 1310 \textit{Predictability in evolution} project C03. We gratefully acknowledge Polish high-performance computing infrastructure PLGrid (HPC Centers: ACK Cyfronet AGH) for providing computer facilities and support within computational grant no. PLG/2024/017221.

{
    \small
    \bibliographystyle{ieeenat_fullname}
    \bibliography{main}
}

\clearpage
\appendix
\setcounter{section}{0}
\renewcommand{\thesection}{\arabic{section}}

\twocolumn[
\begin{center}
    {\Large \bfseries LumiMotion: Improving Gaussian Relighting with Scene Dynamics\par}
    \vspace{0.25em}
    {\Large Supplementary Material\par}
    \vspace{2.0em}
\end{center}
]

\section{Code and data repository}
Code and data are included in our repository:  

\url{https://github.com/joaxkal/LumiMotion}.

\section{Additional videos and figures}

\subsection{Videos}
Please refer to our attached videos for more results on:
\begin{itemize}
    \item \textbf{ENeRF dataset \cite{enerf} }: real world data, moving actors with wall background. Actors cast strong shadow.
    \item \textbf{DNA dataset \cite{dna}}: real world multiview data, moving actors with additional items like stool, table, hair dryer. 
    \item \textbf{our synthetic scenes.}
\end{itemize}

\subsection{Figures}
\begin{figure*}[]
\centering
\includegraphics[width=0.9\linewidth]{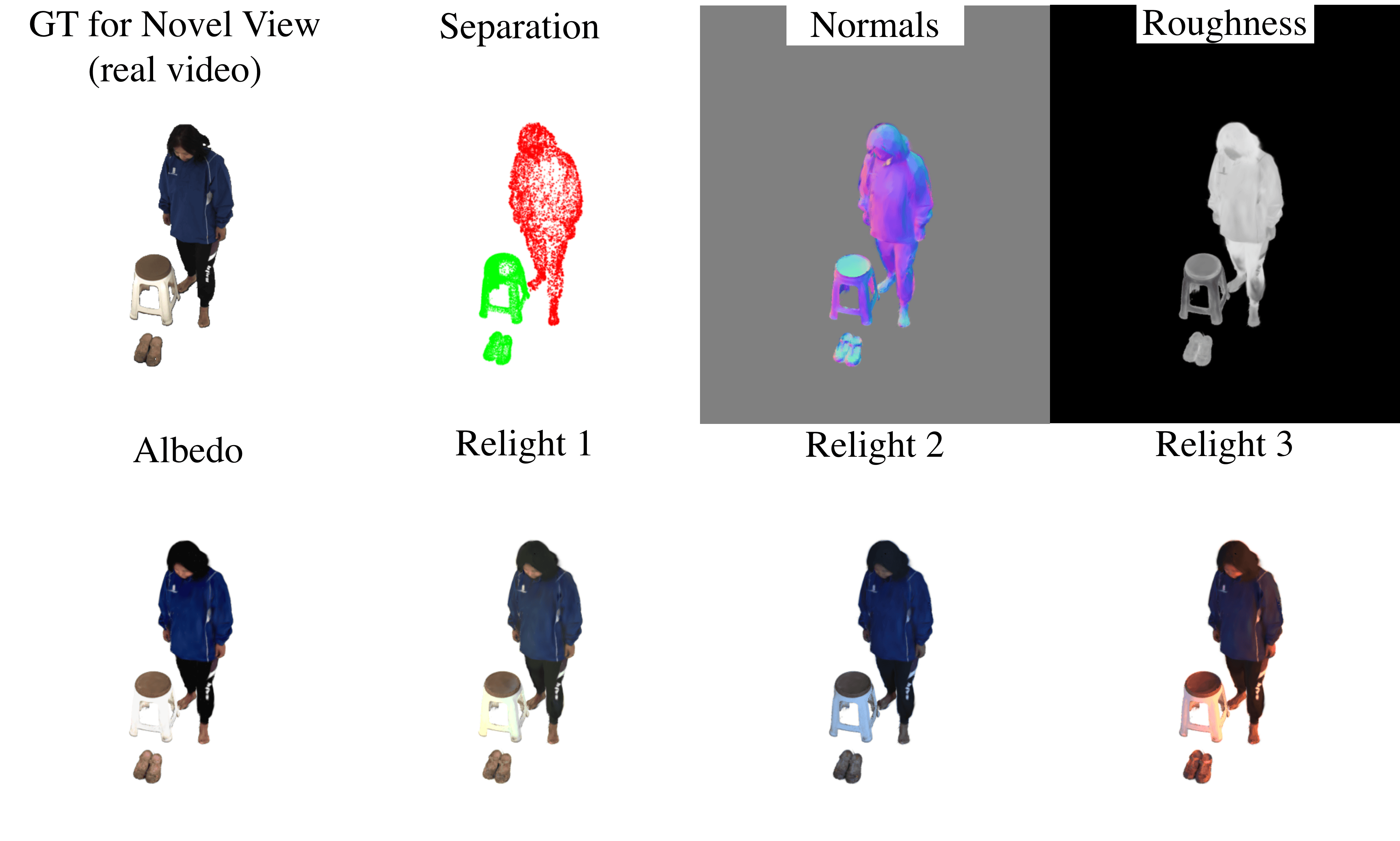}
\caption{DNA - Actor going to sit on a stool. Please see attached video for this scene.}
\label{fig:dna_shoes}
\vspace{-0.0cm}
\end{figure*}

\begin{figure*}[]
\centering
\includegraphics[width=0.9\linewidth]{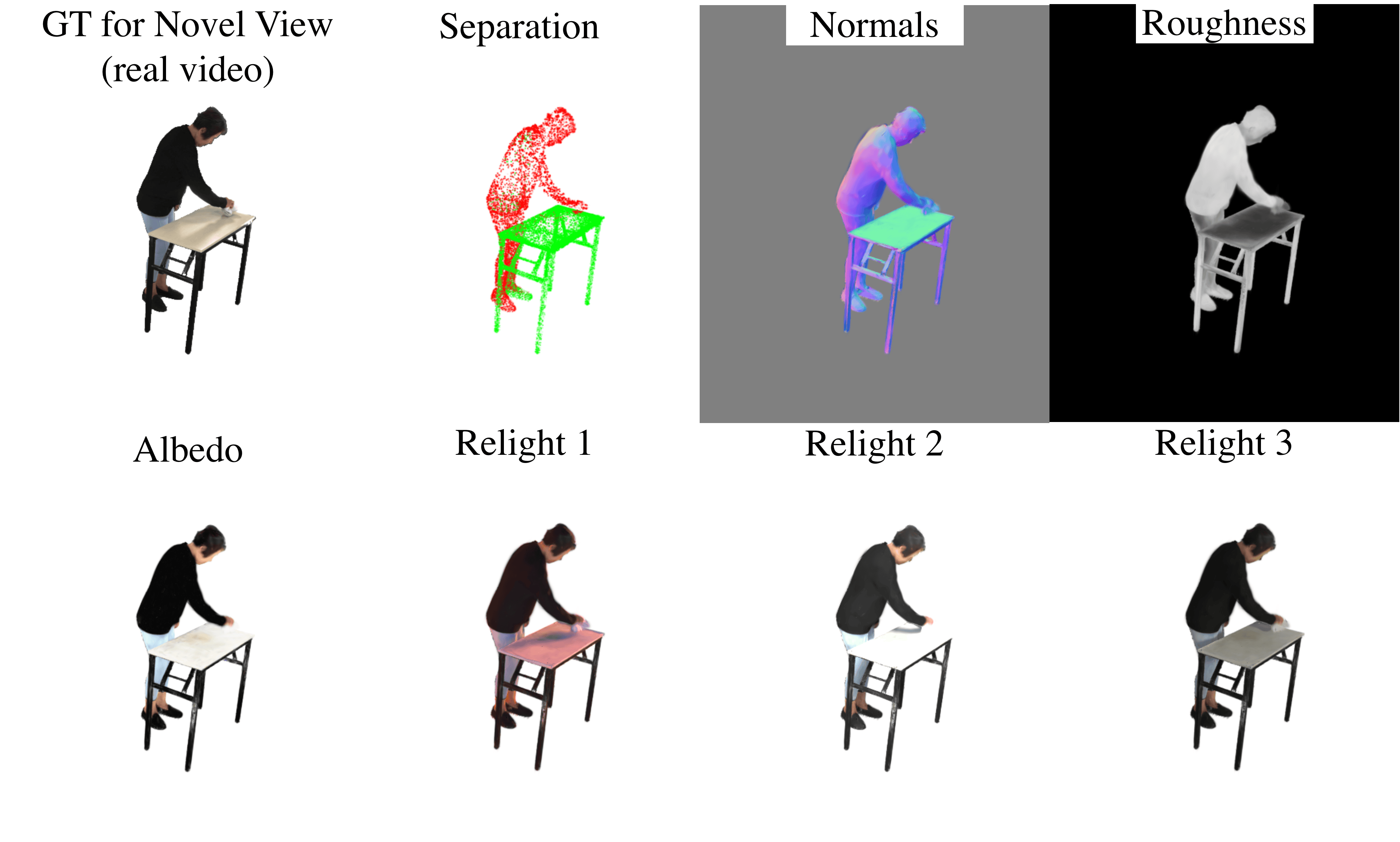}
\caption{DNA – Actor cleaning a specular table.
Although the actor’s arm casts a strong shadow on the tabletop, the Gaussians on the table are correctly classified as static. Please see attached video for this scene.}
\label{fig:dna_table}
\vspace{-0.0cm}
\end{figure*}

\begin{figure*}[]
\centering
\includegraphics[width=0.9\linewidth]{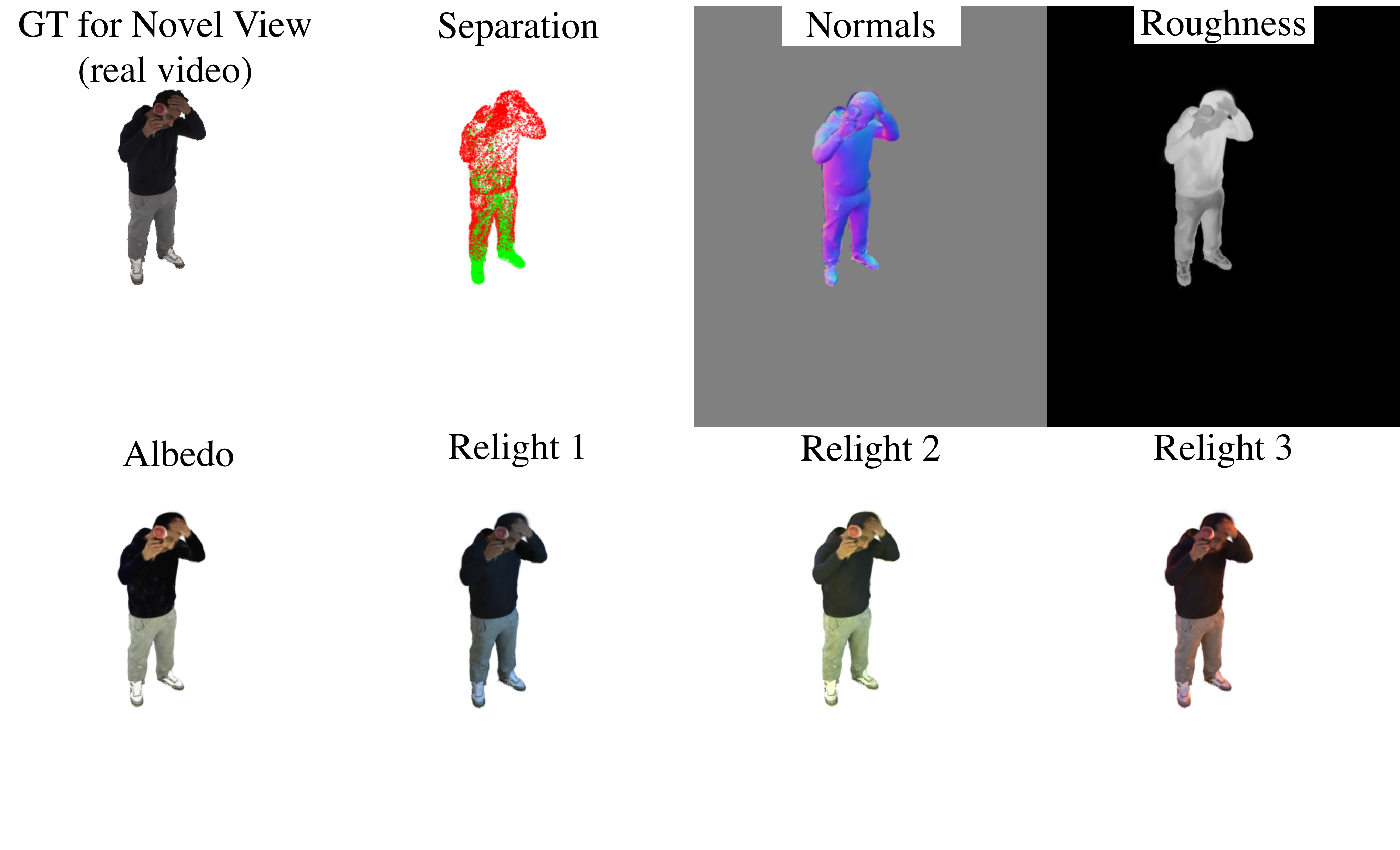}
\caption{DNA - Actor with a hair dryer. Actor's feet are static and only his upper body parts are moving. Please see attached video for this scene.}
\label{fig:dna_hairdryer}
\vspace{-0.0cm}
\end{figure*}

We present additional renders for
\begin{itemize}
\item \textbf{DNA dataset \cite{dna}}: moving humans with various items (table, stool, hair dryer) in Fig.~\ref{fig:dna_table}, \ref{fig:dna_shoes}, \ref{fig:dna_hairdryer}.
\item \textbf{more comparisons with baseline methods} in Fig.~\ref{fig:more_comp_hook}, \ref{fig:more_comp_mouse}, \ref{fig:more_comp_standup}.
\end{itemize}

\section{Extended results}
In Tab.~\ref{tab:app_quantitative_eval_nosmooth} we present extended results, including Novel View Synthesis (NVS) and Roughness. 

\subsection{Novel View Synthesis}
We show that \textbf{the dynamic setting we use is significantly more challenging than the static setup for baselines}, as reflected in the novel view synthesis metrics. Despite this, \ours{} achieves strong results for materials and relighting, demonstrating the effectiveness of our approach.

\begin{figure*}[b!]
\centering
\includegraphics[width=0.99\linewidth]{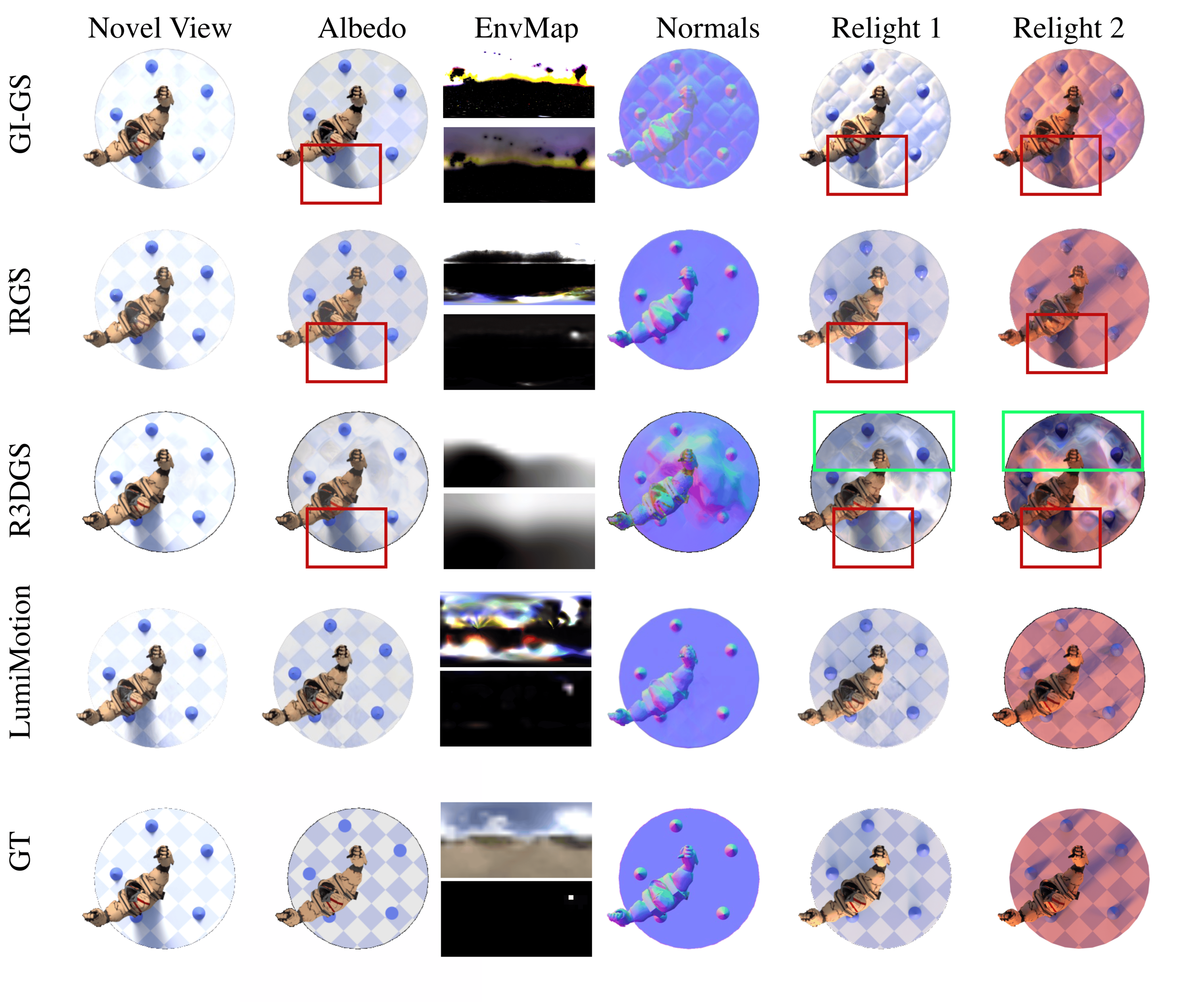}
\caption{Scene 1. Our method produces much clearer albedo than the baselines. Although we train on dynamic scenes—more challenging than the static ones used by baselines—our results still show smooth and accurate geometry and normals. The presented environment map (top shows the map truncated to $[0, 1]$ while bottom shows it scaled so that the maximum value is $1$) captures the main light direction more accurately than environment maps estimated from static baselines. We mark in color boxes fine details like artifacts including baked-in shadows, which clearly show superiority of our method. We highlight fine details—such as artifacts and baked-in shadows—in color boxes, clearly demonstrating the superiority of our method.}
\label{fig:more_comp_hook}
\vspace{-0.0cm}
\end{figure*}

\begin{figure*}[h!]
\centering
\includegraphics[width=0.99\linewidth]{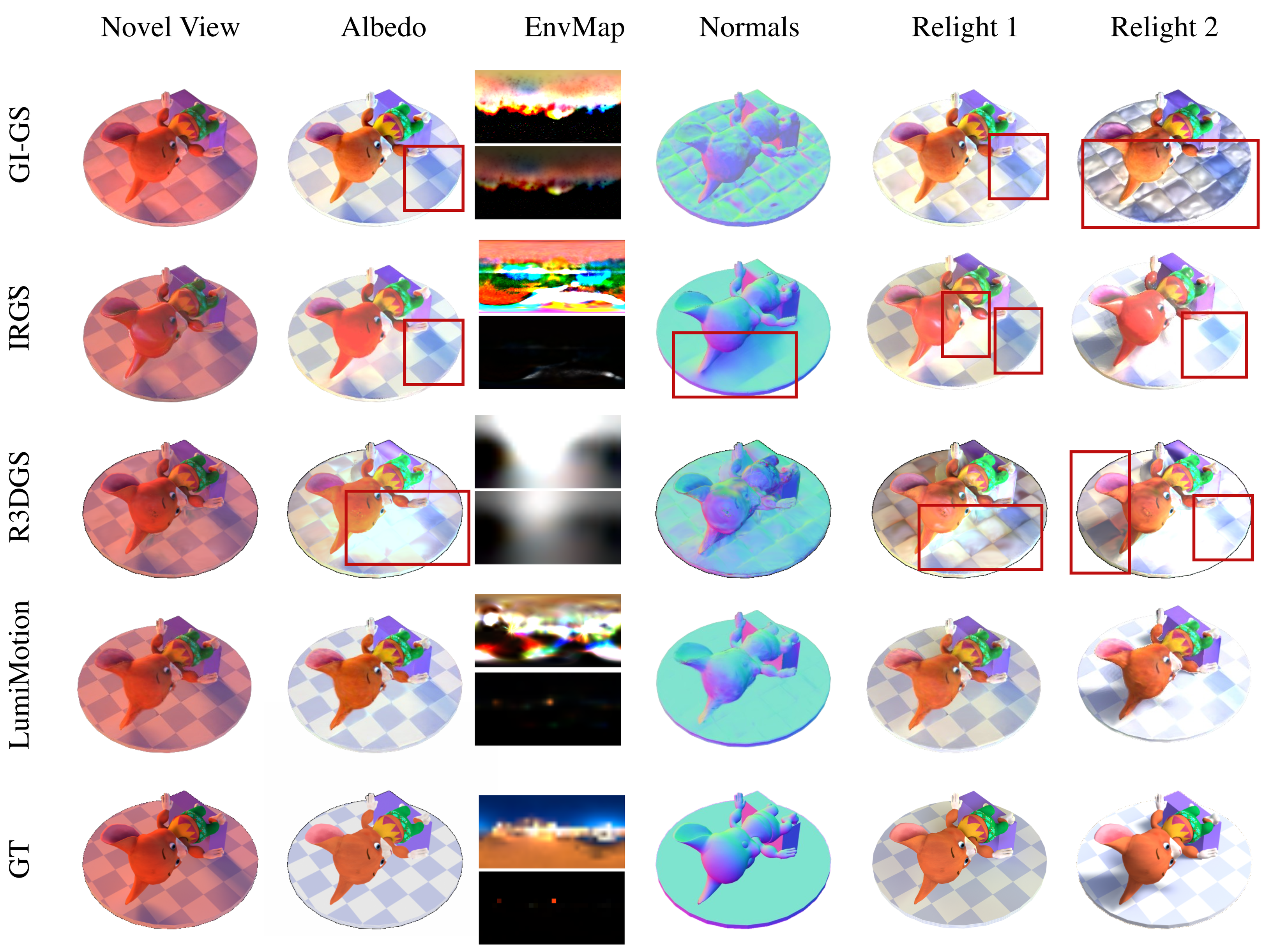}
\caption{Scene 2. Our method produces much clearer albedo than the baselines. Although we train on dynamic scenes—more challenging than the static ones used by baselines—our results still show smooth and accurate geometry and normals. The presented environment map (top shows the map truncated to $[0, 1]$, while bottom shows it scaled so that the maximum value is $1$) captures the main light direction more accurately than environment maps estimated from static baselines. We highlight fine details—such as artifacts and baked-in shadows—in color boxes, clearly demonstrating the superiority of our method.}
\label{fig:more_comp_mouse}
\vspace{-0.0cm}
\end{figure*}

\begin{figure*}[h!]
\centering
\includegraphics[width=0.99\linewidth]{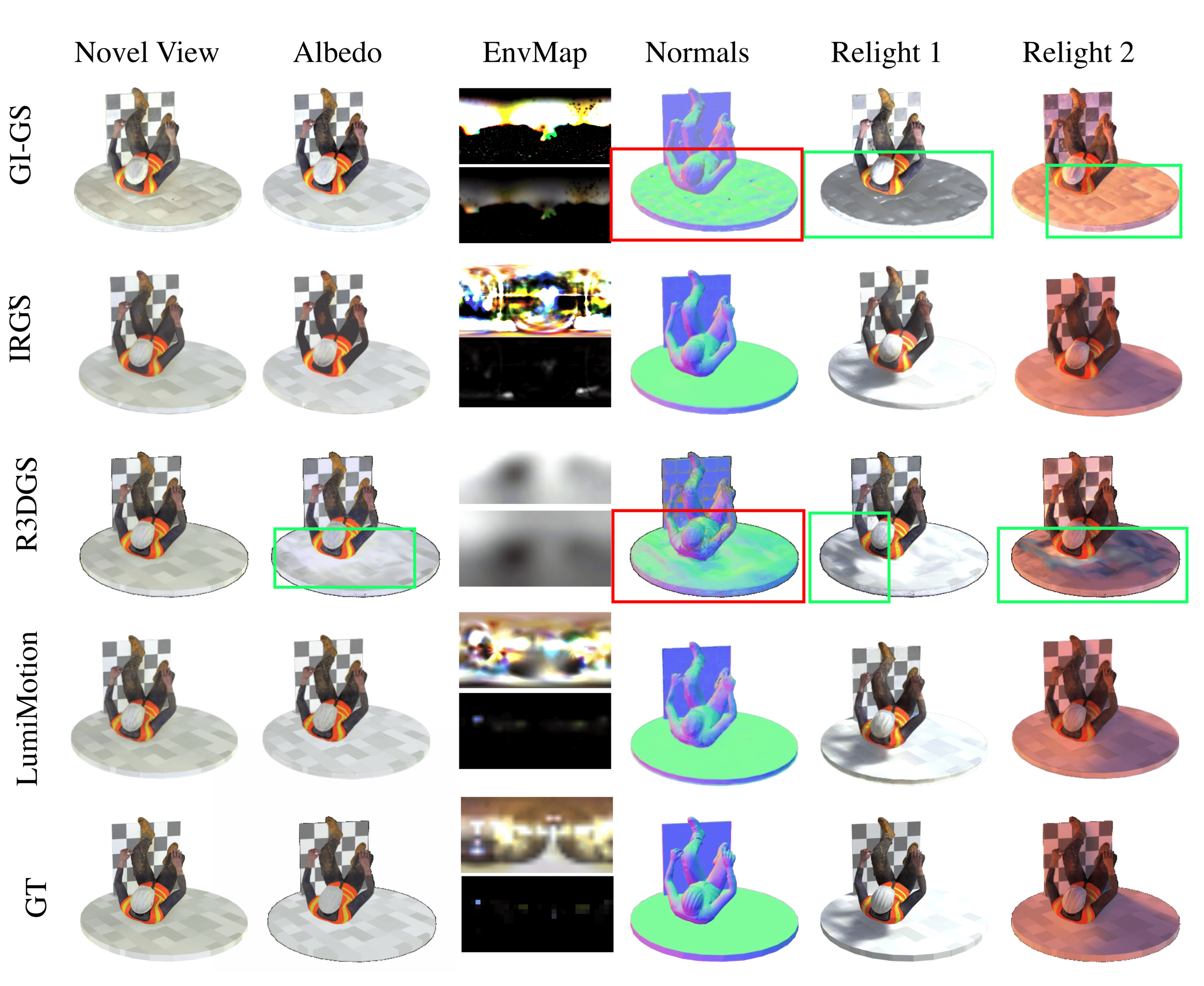}
\caption{Scene 3. This scene was captured under mild training illumination, with no strong shadows present. As a result, the static baselines exhibit relatively well-separated albedo in the training image (no baked in shadows). We demonstrate that in this case our method achieves relighting and geometry quality competitive with the latest state-of-the-art IRGS, despite IRGS being a static baseline. The presented environment map (top shows the map truncated to $[0, 1]$, while bottom shows is scaled so that the maximum value is $1$) captures the main light direction more accurately than environment maps estimated from static baselines. }
\label{fig:more_comp_standup}
\vspace{-0.0cm}
\end{figure*}

Please note that the high NVS scores of static baselines are also caused by \textbf{overfitting} to the training lighting conditions. When evaluated under novel illumination, their performance drops significantly, which is also consistent with our qualitative observations (Fig.~\ref{fig:more_comp_hook}, \ref{fig:more_comp_mouse}, \ref{fig:more_comp_standup}) . For clarity, we report the PSNR drop in the last column of the table.   

\textbf{This highlights the effectiveness of our separation strategy and the consistent behavior of our method across both training and novel lighting conditions.}

\subsection{Roughness} 
\begin{figure*}[]
\centering
\includegraphics[width=0.9\linewidth]{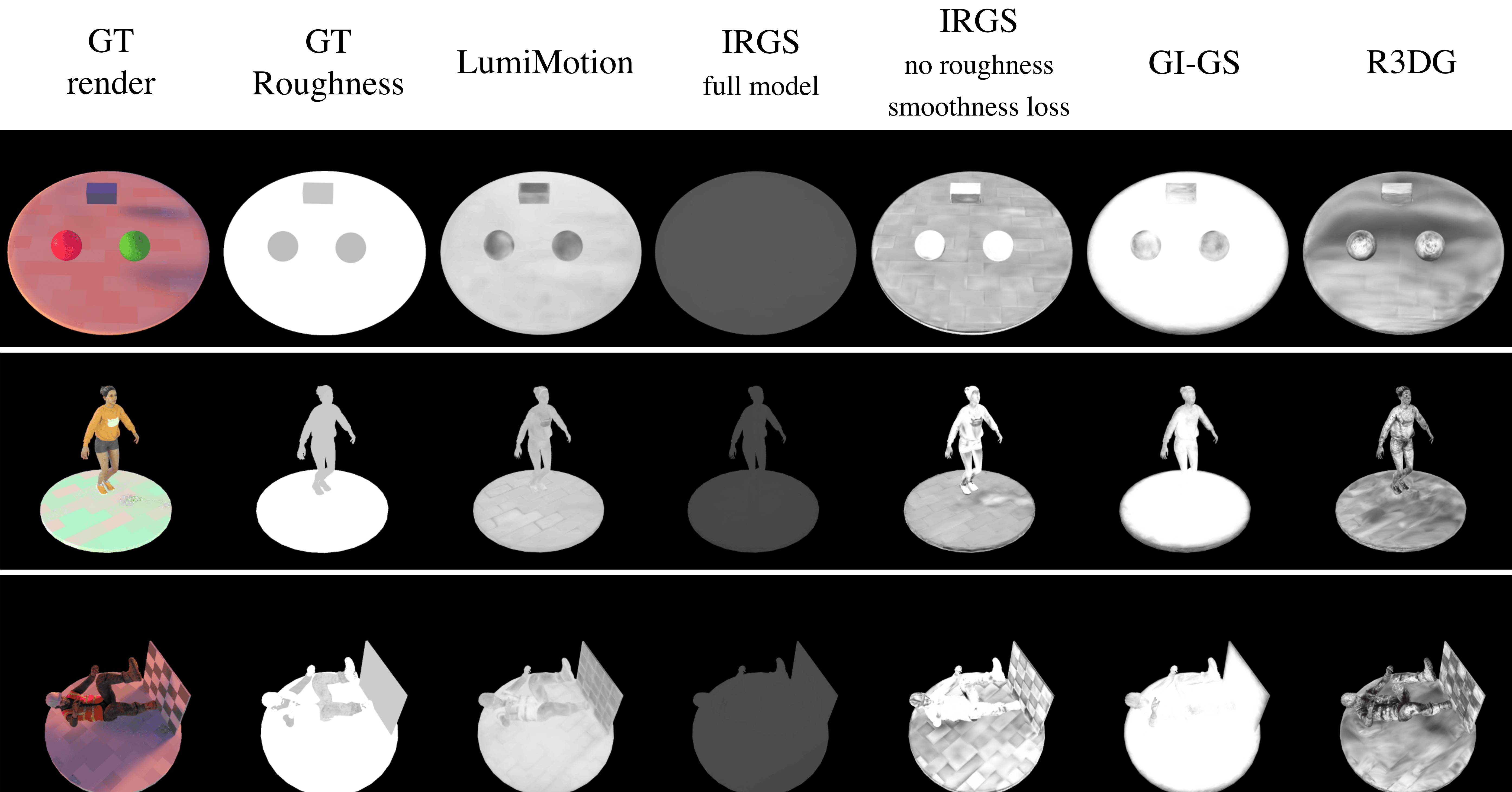}
\caption{Roughness estimation. Baseline methods fail to recover realistic roughness maps. For IRGS, the standard loss set yields roughness values with little to no surface detail (see the `full model` column). Removing smoothness constraints causes different issues, for example: in the top row, the plate is estimated as specular and the items as rough, whereas the ground truth shows the opposite. Other baselines also fail to disentangle material properties and estimate roughness reliably. In contrast, \ours~produces smooth and largely accurate roughness maps.}
\label{fig:roughness_comparison}
\vspace{-0.0cm}
\end{figure*}

We present additional results for roughness estimation. For fair comparison, we experimented with modifying the default IRGS configuration. We found out its standard 
 smooth constraint on roughness adversely affects material estimation - produces roughness maps that are unnaturally smooth and lack detail. See Fig.~\ref{fig:roughness_comparison} for comparison. Therefore, in the table we also present results obtained by training IRGS without this loss term.   
 
 \textbf{Please note that \ours{} consistently achieves significantly lower MSE for roughness comparing to even the closest baseline, IRGS}. 

\begin{table*}[]
\centering
\small
\setlength{\tabcolsep}{2.5pt}
\renewcommand{\arraystretch}{1.1}
\caption{\textbf{\ours{} for novel view synthesis, material estimation, and relighting.} Static methods use only one timestep, while \ours{} operates on a dynamic scene. Testing is performed on the same views and timestep as in the static setting. Results are grouped by train-test lighting conditions. For IRGS, we denote training scheme without smooth constraint on roughness by $\dag$. 
\newline
\newline
\textbf{NVS}: Note that \textbf{the dynamic setting we use is significantly more challenging than static setup for baselines}. Despite this, \ours{} achieves strong results for albedo and relight, demonstrating the effectiveness of our approach. Please note that the high NVS scores of static baselines are also caused by \textbf{overfitting} to the training lighting conditions. When evaluated under novel illumination, their performance drops significantly. This observation is consistent with our qualitative results. We show PSNR drop in the last column. \textbf{This highlights the effectiveness of our separation strategy and the consistent behavior of our method across both train and test lighting.}  
\newline
\newline
\textbf{Material}: We achieve significantly better material estimation than the closest baseline, IRGS, regardless of its training setup. Notably, \ours{} consistently produces \textbf{higher-quality albedo} and achieves at least a \textbf{2× lower roughness MSE} compared to IRGS.}
\begin{tabular}{l@{\hskip 12pt}ccc@{\hskip 12pt}ccc@{\hskip 12pt}c@{\hskip 12pt}ccc@{\hskip 12pt}c}
\hline
\textbf{Method}
& \multicolumn{3}{c@{\hskip 12pt}}{\textbf{Novel View Synthesis}} 
& \multicolumn{3}{c@{\hskip 12pt}}{\textbf{Albedo}} 
& \multicolumn{1}{c@{\hskip 12pt}}{\textbf{Roughness}}
& \multicolumn{3}{c@{\hskip 12pt}}{\textbf{Relight}} 
& \multirow{2}{*}{\shortstack{$\Delta$PSNR\\NVS$\rightarrow$Relight}} \\
& PSNR $\uparrow$ & SSIM $\uparrow$ & LPIPS $\downarrow$
& PSNR $\uparrow$ & SSIM $\uparrow$ & LPIPS $\downarrow$
& MSE $\downarrow$
& PSNR $\uparrow$ & SSIM $\uparrow$ & LPIPS $\downarrow$
& \\
\hline
\multicolumn{12}{c}{\textbf{Dam Wall $\rightarrow$ Harbour Sunset}} \\
\hline
R-3DGS  & $35.031$  & $0.987$ & $0.035$ & $20.744$ & $0.900$ & $0.128$ & $0.066\pm0.005$ & $21.220$ & $0.915$ & $0.112$ & \cellcolor{red!20}{$39.5\%$} \\
GI-GS   & $26.749$  & $0.956$ & $0.066$ & $20.943$ & $0.906$ & $0.105$ & $0.036\pm0.001$ & $18.431$ & $0.868$ & $0.139$ & \cellcolor{red!20}{$31.1\%$} \\
IR-GS   & $32.207$  & $0.983$ & $0.021$ & $22.888$ & $0.936$ & $0.076$ & $0.136 \pm 0.040$ & $26.177$ & $0.953$ & $0.064$  & \cellcolor{yellow!20}{$18.7\%$} \\
IR-GS$\dag$ & $32.639$  & $0.985$ & $0.019$ & $23.512$ & $0.935$ & $0.080$ & $0.024\pm0.010$ & $27.156$ & $0.954$ & $0.067$ & \cellcolor{yellow!20}{$16.8\%$} \\
\ours{} & $26.948$  & $0.952$ & $0.025$ & \boldmath $27.268$ & \boldmath $0.952$ & \boldmath $0.069$ & \boldmath $0.012\pm0.002$& $26.037$ & $0.928$ & $0.060$ & \cellcolor{green!20}{\boldmath $3.4\%$} \\
\hline
\multicolumn{12}{c}{\textbf{Chapel Day → Golden Bay}} \\
\hline
R-3DGS  & $36.986$ & $0.989$ & $0.028$ & $22.463$ & $0.927$ & $0.096$ & $0.044\pm0.008$ & $22.282$ & $0.943$ & $0.081$ & \cellcolor{red!20}{$39.8\%$} \\
GI-GS   & $29.489$ & $0.971$ & $0.057$ & $24.733$ & $0.955$ & $0.056$ & $0.031\pm0.002$ & $22.673$ & $0.880$ & $0.125$ & \cellcolor{red!20}{$23.1\%$} \\
IR-GS   & $33.580$ & $0.983$ & $0.022$ & $23.769$ & $0.956$ & $0.053$ & $0.128\pm0.045$ & $28.157$ & $0.966$ & $0.046$ & \cellcolor{yellow!20}{$16.2\%$} \\
IR-GS $\dag$  & $34.212$ & $0.985$ & $0.019$ & $24.085$ & $0.956$ & $0.052$ & $0.028\pm0.012$ & $28.702$ & $0.968$ & $0.046$ & \cellcolor{yellow!20}{$16.1\%$} \\
\ours{} & $27.636$ & $0.952$ & $0.022$ & \boldmath $30.838$ & \boldmath $0.973$ & \boldmath $0.036$ & \boldmath $0.011\pm0.002$ &$28.563$ & $0.939$ & $0.041$ & \cellcolor{green!20}{\boldmath $-3.3\%$} \\
\hline
\multicolumn{12}{c}{\textbf{Golden Bay → Dam Wall}} \\
\hline
R-3DGS  & $36.096$ & $0.988$ & $0.028$ & $19.945$ & $0.899$ & $0.133$ & $0.039\pm0.010$ & $19.563$ & $0.918$ & $0.118$ & \cellcolor{red!20}{$45.8\%$} \\
GI-GS   & $34.402$ & $0.982$ & $0.031$ & $21.295$ & $0.932$ & $0.087$ &$0.031\pm0.003$ & $17.636$ & $0.823$ & $0.132$ & \cellcolor{red!20}{$48.8\%$} \\
IR-GS   & $34.404$ & $0.980$ & $0.026$ & $20.910$ & $0.937$ & $0.082$ & $0.145\pm0.027$ & $25.009$ & $0.955$ & $0.060$ & \cellcolor{red!20}{$27.3\%$} \\
IR-GS$\dag$  & $35.978$ & $0.985$ & $0.020$ & $21.199$ & $0.936$ & $0.081$ & $0.021\pm0.008$ & $25.252$ & $0.957$ & $0.058$ & \cellcolor{red!20}{$29.8\%$} \\
\ours{} & $29.859$ & $0.954$ & $0.023$ & \boldmath $27.929$ & \boldmath $0.959$ & \boldmath $0.058$ & \boldmath $0.010\pm0.002$& $25.405$ & $0.936$ & $0.048$ & \cellcolor{yellow!20}{\boldmath $14.9\%$} \\
\hline
\end{tabular}
\label{tab:app_quantitative_eval_nosmooth}
\end{table*}

\FloatBarrier

\clearpage
\twocolumn[
\vspace*{0.1cm}]
\section{Separation - additional example of ablation and hyperparameter influence}

\begin{figure}[!h]
\centering
\includegraphics[width=1.0\linewidth]{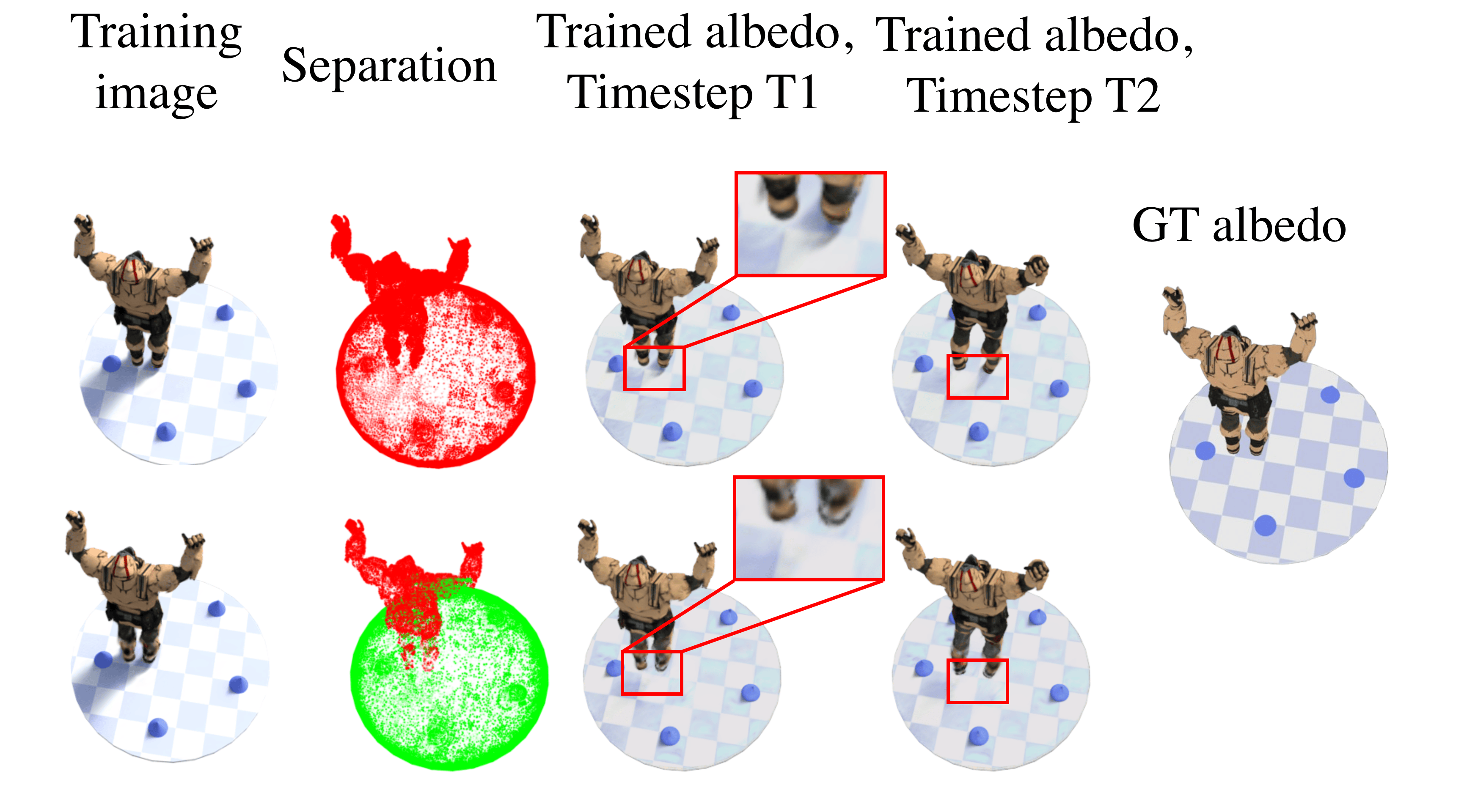}
\caption{Separation - ablation study.  \textbf{Please zoom in for details.}}
\label{fig:separation_abl_hook}
\vspace{-0.0cm}
\end{figure}

\begin{figure}[!h]
\centering
\includegraphics[width=1.0\linewidth]{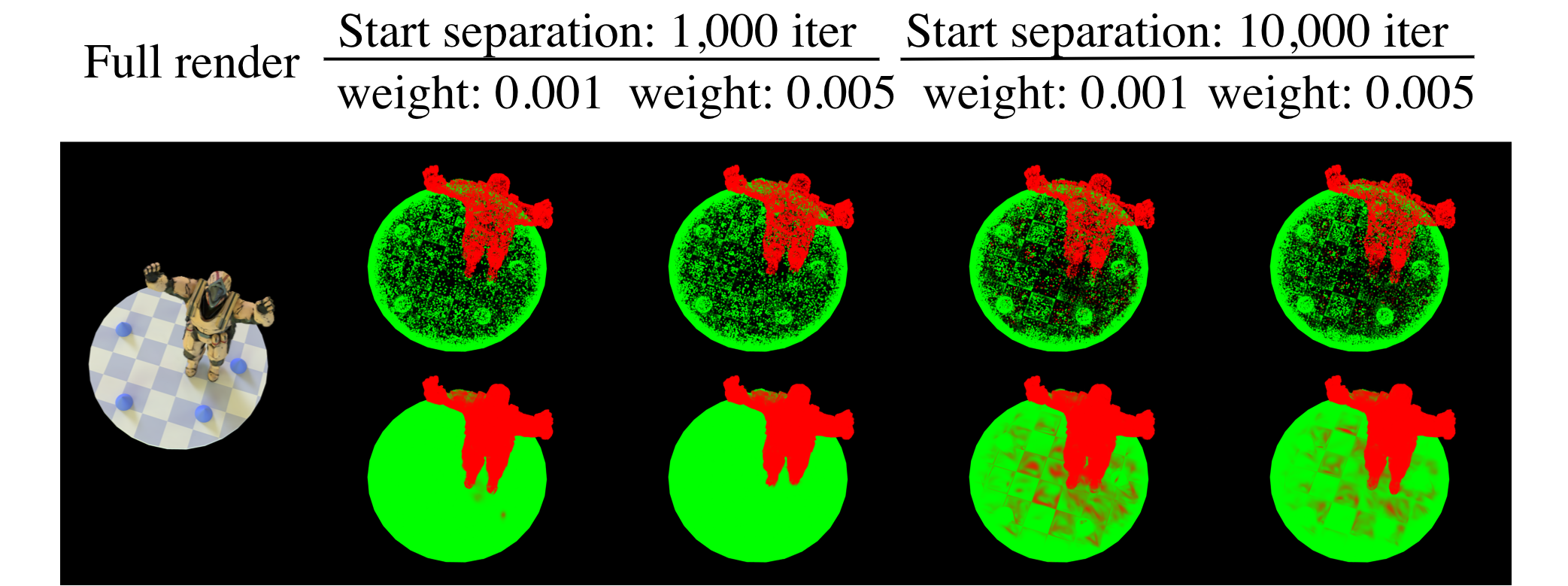}
\caption{Separation – impact of hyperparameters. Top: Renderings using downsized Gaussians illustrate uneven Gaussian placement on the plate when the separation loss is introduced too late, in 10,000th iteration. In thia case several Gaussians on the plate are incorrectly classified as dynamic and move with the shadows. Bottom: Renderings with full-sized Gaussians highlight the extent of this misclassification. \textbf{Please zoom in for details.}}
\label{fig:separation_weight_hook}
\vspace{-0.0cm}
\end{figure}

In Fig.~\ref{fig:separation_weight_hook}, we illustrate the influence of separation hyperparameters. Our separation method robustly detects moving parts of jumping actor. Depending on the scene, a delayed start or a separation value that is too low may impair the penalization of static regions. In Fig.~\ref{fig:separation_abl_hook} we show that without separation, strong moving shadow on the plate is modeled by moving Gaussians. Our separation strategy allows for cleaner albedo without shadow artifacts.

\FloatBarrier
\clearpage
\twocolumn[
\vspace*{0.1cm}
]
\section{Our dataset}
\begin{figure}[!b]
\centering
\includegraphics[width=0.99\linewidth]{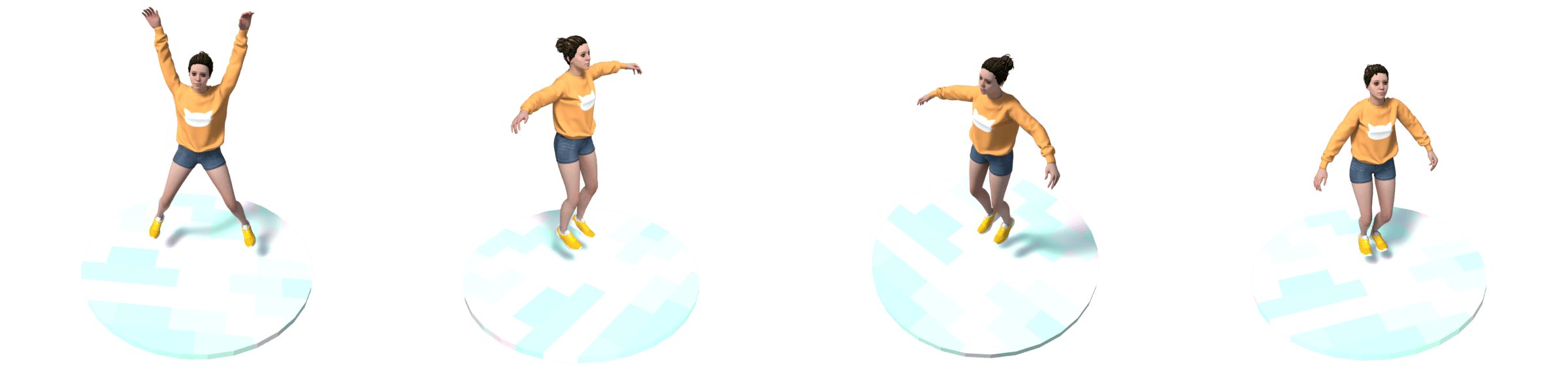}
\caption{Example visualization of the `jumpingjacks` scene at time steps 6, 46, 53, and 86. Note that each time step corresponds to a different camera pose, so the renderings are shown from varying viewpoints.}
\label{fig:app_timesteps}
\vspace{-0.0cm}
\end{figure}

We provide additional details about our synthetic dataset and its generation process. We build 5 synthetic datasets in Blender, using Mixamo\footnote{https://www.mixamo.com/} platform and simple Blender meshes. We prepared each scene in two versions: dynamic and static. For dynamic version, we use D-NeRF \cite{pumarola2021d} like setup with different camera view for each timestep, creating a multi-view, dynamic scene suitable for evaluating relighting and novel view synthesis (see Fig.~\ref{fig:app_timesteps}). For static variant, we use the same camera views and only one timestep. All scenes but `spheres` contain 150 frames, for `spheres` there are 100 frames.

\begin{figure}[!b]
\centering
\includegraphics[width=0.99\linewidth]{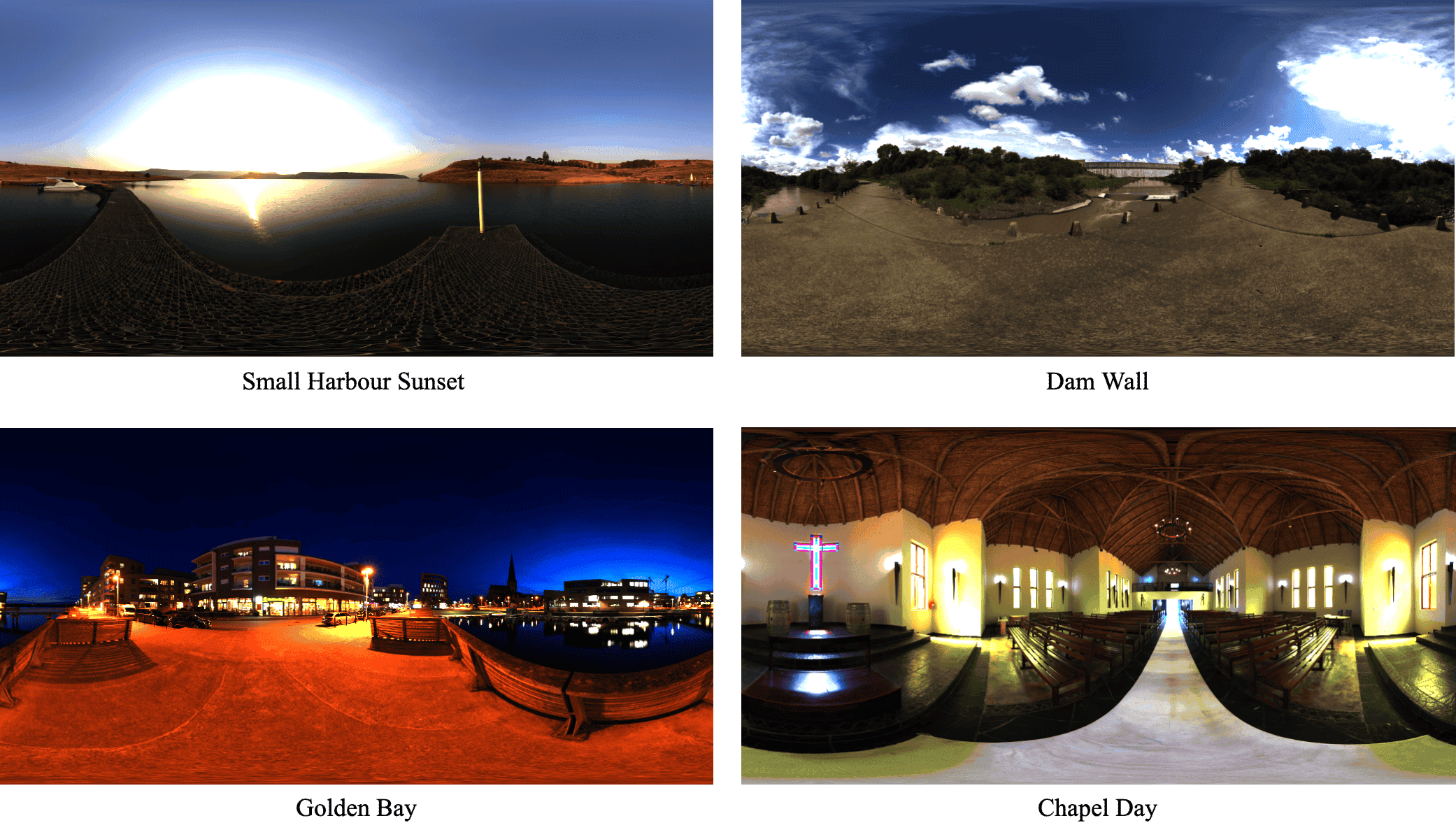}
\caption{Environment maps used to light our scenes.}
\label{fig:app_envmaps}
\vspace{-0.3cm}
\end{figure}

Each scene is relit using four high-dynamic-range (HDR) environment maps from PolyHaven\footnote{https://polyhaven.com/hdris} selected to span diverse lighting conditions (the environment maps are rotated by us such that the dominant light source appears from various directions):\\
\textbf{•} Small Harbour Sunset\\
\textbf{•} Dam Wall\\
\textbf{•} Golden Bay\\
\textbf{•} Chapel Day. \\
We show all environment maps in Fig.~\ref{fig:app_envmaps}.

\begin{figure*}[h!]
\centering
\includegraphics[width=0.6\linewidth]{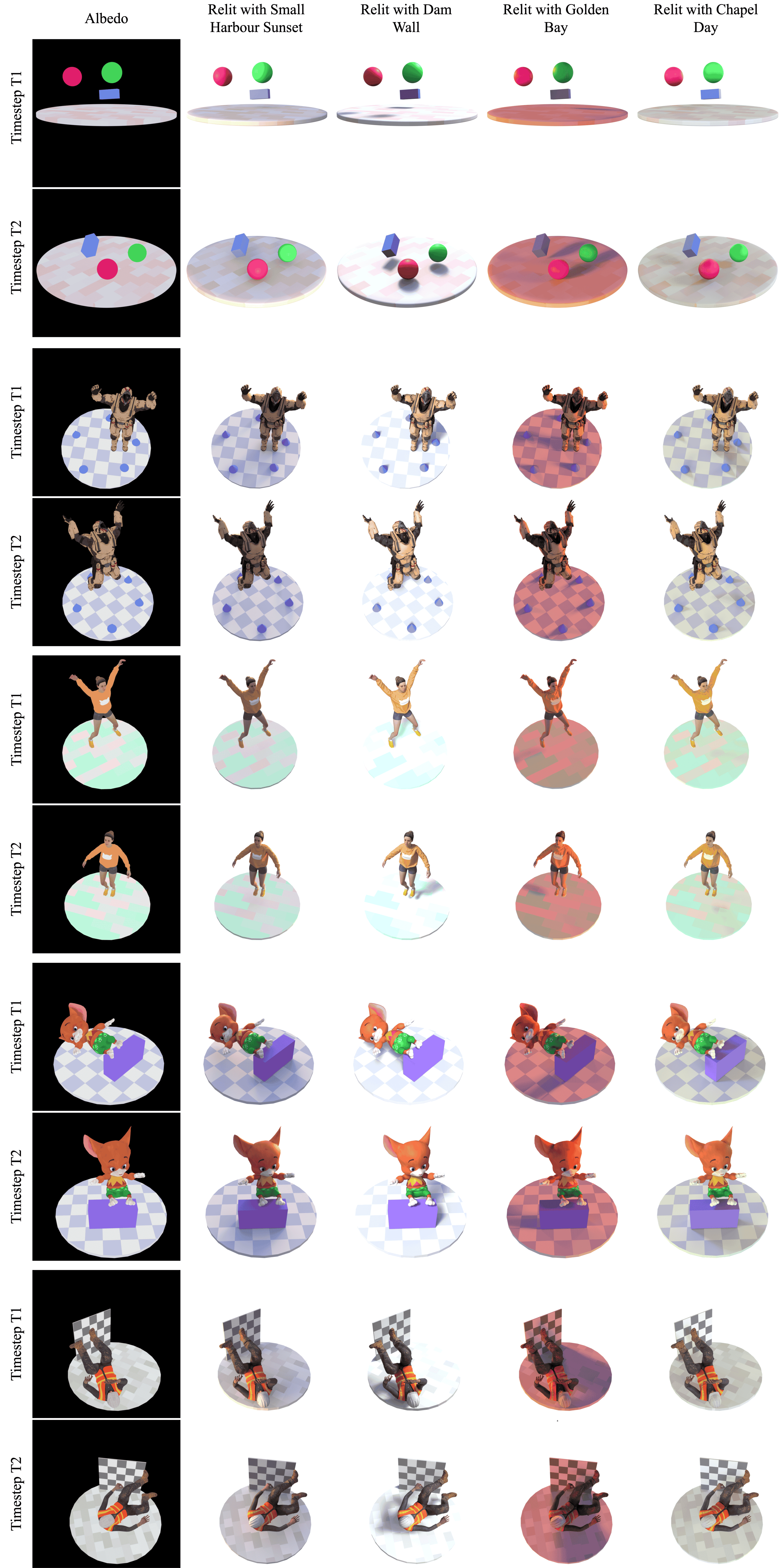}
\caption{Example ground truth renders of our scenes.}
\label{fig:app_comp}
\vspace{-0.0cm}
\end{figure*}

Originally the environment maps are in 4K resolution; we rescale them to $32 \times 16$ to introduce blur and avoid extremely sharp shadows. To enable shadow analysis, each scene is composed of both dynamic and static elements (e.g., plates and blocks), all of which cast shadows. For each dataset, we provide ground truth albedo. The amount of specular reflectance varies across scenes and objects. Example renders from our dataset are shown in Fig.~\ref{fig:app_comp}.
\vspace{-0.0cm}

\FloatBarrier

\newpage

\section{Implementation details}
We train each scene in two stages: 35{,}000 iterations in Stage 1 and 20{,}000 iterations in Stage 2. Our MLP architecture follows the design proposed in \cite{Yang_2024_CVPR_deformable_gs}, consisting of an 8-layer MLP with a width of 256 units per layer. The learning rate for the MLP is set to 0.0008 and decays exponentially to 0.00008.

In Stage 1, we train using a combination of loss terms with the following weights (brackets show hyperparameter search range:
\begin{equation}
\begin{split}
\lambda_n = 0.002, \quad \lambda_d = 1000, \quad \lambda_o = 0.1, \\
\lambda_P = \{0.001, 0.005\}, \quad \lambda_{\Delta c} = 0.01, \quad \lambda_{\Delta \mu} = \{0.0, 0.001\}
\end{split}
\end{equation}.

In Stage 2, we optimize the albedo, which is an RGB value assigned to each Gaussian, and roughness - values constant over time, and the environment map. The learning rates for the environment map, albedo and roughness are set to 0.2, 0.01, 0.005 respectively. The training environment map has a resolution of $32 \times 16$ for synthetic data and $128 \times 64$ for DNA scenes and ENERF data considering its very sharp shadows. We also finetune Gaussian colors from Stage 1 together with MLP head responsible for modeling $d_{color}$. This is important, since we use Stage 1 colors to compute indirect light for training, following \cite{irgs}. We also finetune opacity to allow the model to remove some relight-related artifacts visible during training. The remaining parameters and MLP parts are frozen so the learned geometry from the Stage 1 is remained. All finetuned parameters in Stage 2 have their original lr lowered by 10 times.

During synthetic training, we sample 512 from the environment map. We randomly select $N_r = 2^{18}$ rays per iteration, resulting in $2^{18} / 512$ pixels used to compute the $\ell_1$ loss for synthetic data. For ENERF we use 1024 samples and $2^{18}\cdot16$ rays.

At inference time, we relight scenes using 1024 or 2048 sampled rays.  

Please refer to our repository for the exact hyperparameter settings to reproduce our results.

\section{Limitation - example}
\begin{figure}[t!]
\centering
\includegraphics[width=0.99\linewidth]{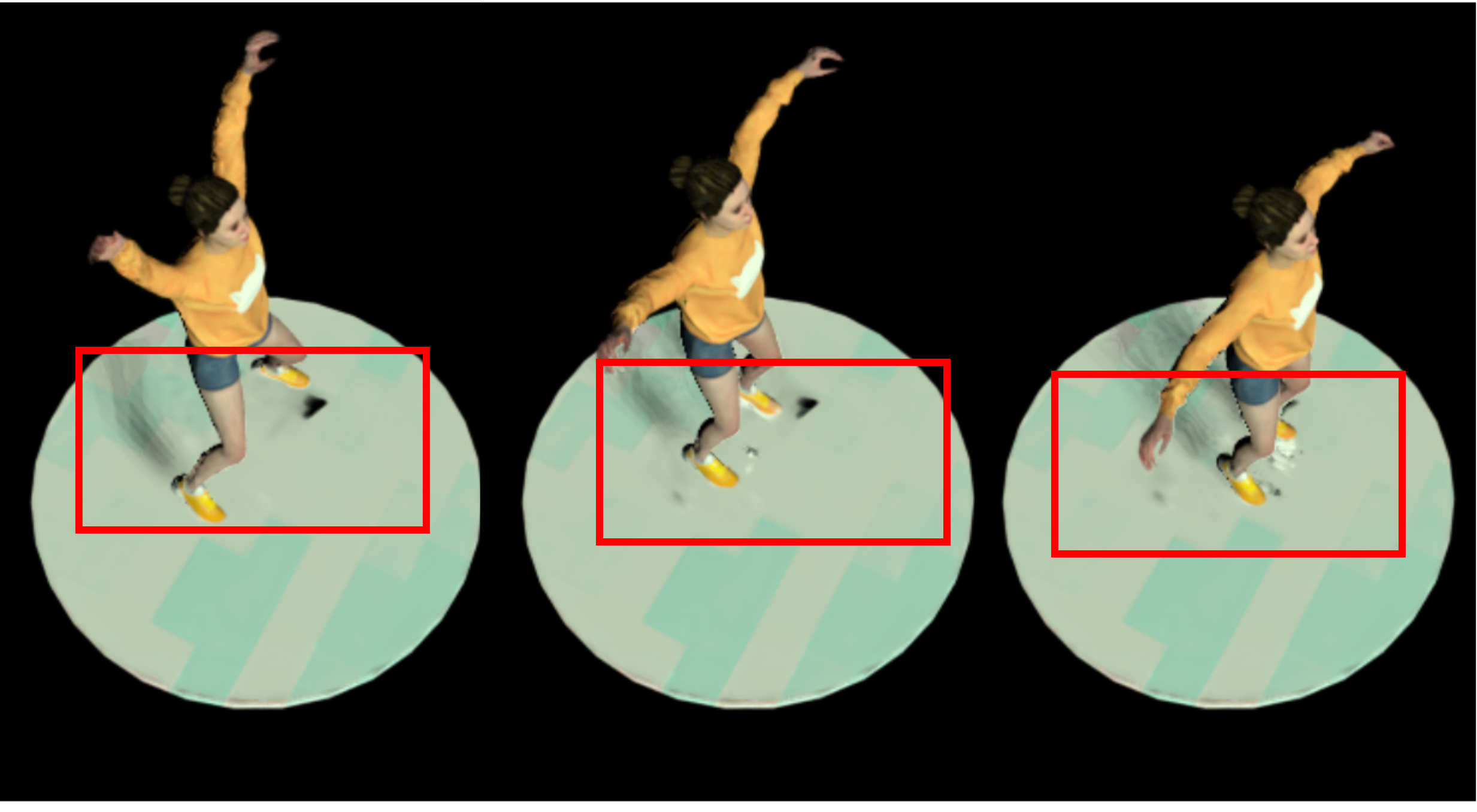}
\caption{Limitations of simple separation strategy. Some Gaussians between the plate surface and the shoes are neither part of the static plate nor clearly part of the dynamic shoe.}
\label{fig:app_fails}
\end{figure}

In Fig.~\ref{fig:app_fails}, we illustrate the limitations of our dynamic training strategy. For more complex and detailed motions, for example near surfaces, simple separation may need to be replaced with more specialized supervision, such as optical flow.

\section{Full affiliations}
The full affiliations, abbreviated in the author section due to space constraints, are as follows: (1) Warsaw University of Technology, Poland; (2) Sano Centre for Computational Medicine, Kraków, Poland; (3) Institute for Biomedical Informatics, Faculty of Medicine and University Hospital Cologne, University of Cologne, Germany; (4) Faculty of Mathematics and Natural Sciences, University of Cologne, Germany; (5) Center for Molecular Medicine Cologne (CMMC), Faculty of Medicine and University Hospital Cologne, University of Cologne, Germany; (6) Jagiellonian University, Kraków, Poland; (7) IDEAS Research Institute, Warsaw, Poland; (8) Microsoft.

\end{document}